\pdfoutput=1
\documentclass{article}

\usepackage{microtype}
\usepackage{graphicx}
\usepackage{booktabs}
\usepackage{hyperref}

\usepackage{amsmath}
\usepackage{amssymb}
\usepackage{amsthm}
\newtheorem{proposition}{Proposition}
\usepackage{xspace}
\usepackage{xcolor}
\usepackage{colortbl}
\usepackage{multirow}
\usepackage{makecell}
\usepackage{enumitem}
\usepackage[normalem]{ulem}
\usepackage[accepted]{mlsys2025}
\usepackage{cleveref}
\usepackage{xurl}
\usepackage{pgfplots}
\pgfplotsset{compat=1.18}
\usepgfplotslibrary{groupplots}
\usepackage{fitcheck}
\usepackage{algorithm}
\usepackage{algorithmic}
\usepackage{listings}
\usetikzlibrary{fit,positioning,calc,arrows.meta,backgrounds,shapes.misc,decorations.pathreplacing}
\definecolor{hCardD}{HTML}{67101E}
\definecolor{hCard}{RGB}{154,31,31}
\definecolor{hCard3}{HTML}{C8808E}
\definecolor{hCard2}{HTML}{E4C2C8}
\definecolor{hCard1}{HTML}{F6EBED}
\definecolor{hSlate}{HTML}{50595A}
\definecolor{hGrey}{HTML}{8A8B8C}
\definecolor{hGreyL}{HTML}{D6D6D7}
\definecolor{hGrid}{HTML}{B8B8B8}
\definecolor{hTeal}{HTML}{06AEBA}
\definecolor{hTeal2}{HTML}{AFD8DE}
\definecolor{hTeal1}{HTML}{D6EFF2}
\definecolor{hAmber}{HTML}{F8B62D}
\definecolor{hOlive}{HTML}{7E9948}
\colorlet{cGPTQ}{hGrey}
\colorlet{cAWQ}{hTeal}
\colorlet{cParo}{hAmber}
\colorlet{cComp}{hCard}
\colorlet{cQDA}{hCard}
\tikzset{
  hfig/.style={font=\footnotesize, line width=0.6pt, >={Stealth[length=4pt]}},
  hop/.style={draw=black, line width=0.6pt, rounded corners=2pt, align=center, inner xsep=4pt, inner ysep=3pt},
  hgroup/.style={draw=#1, dashed, line width=0.7pt, rounded corners=3pt, inner sep=5pt},
  harrow/.style={->, draw=black, line width=0.7pt},
  hcarrow/.style={->, draw=hCard, line width=0.8pt},
}
\pgfplotsset{hplot/.style={
  font=\footnotesize, tick label style={font=\footnotesize}, label style={font=\footnotesize},
  legend style={font=\footnotesize, draw=none, fill=none},
  axis line style={black, line width=0.6pt}, major tick length=2pt,
  ymajorgrids, grid style={hGrid, line width=0.15pt}}}
\definecolor{hkRed}{RGB}{154,31,31}
\definecolor{hkBlue}{RGB}{122,140,255}
\definecolor{hkCoral}{RGB}{255,137,88}
\definecolor{hkNavy}{RGB}{2,44,69}
\definecolor{hkAmber}{RGB}{234,154,19}
\colorlet{dOrange}{hkCoral!35}
\colorlet{dPurple}{hkBlue!28}
\colorlet{dBlue}{hkBlue!55}
\colorlet{dBlueL}{hkBlue!13}
\colorlet{dGreen}{hkAmber!32}
\colorlet{dYellow}{hkAmber}
\definecolor{dCell}{HTML}{D9D9D9}
\colorlet{dLine}{hkNavy!85!black}
\tikzset{
  dbox/.style={draw=dLine, line width=0.5pt, rounded corners=1.5pt, align=center, inner xsep=3pt, inner ysep=2pt},
  darrow/.style={->, draw=dLine, line width=0.6pt, >={Stealth[length=3.5pt]}},
}
\colorlet{qBanner}{hCard}\colorlet{qBlue}{hTeal1}\colorlet{qGreen}{hTeal1}\colorlet{qOrange}{hCard1}
\colorlet{qPurple}{hCard2}\colorlet{qGray}{hGreyL}\colorlet{qLine}{black}
\tikzset{qfig/.style={hfig}, qbox/.style={hop}, qregion/.style={hgroup=black},
  qbanner/.style={fill=hCard, text=white, font=\footnotesize\bfseries, rounded corners=2pt, inner xsep=5pt, inner ysep=2.5pt},
  qarrow/.style={harrow}}
\pgfplotsset{qplot/.style={hplot}}

\newcommand{\MyPara}[1]{\noindent{\textbf{#1}}~}
\newcommand{\refsec}[1]{\S~\ref{#1}}

\newcommand{\sys}{QDA\xspace}
\newcommand{\syscmd}{qda}

\newcommand{\qthreefive}{Qwen3.5-9B\xspace}
\newcommand{\qthreesix}{Qwen3.6-35B-A3B\xspace}
\newcommand{\qthreeeight}{Qwen3.8-27B\xspace}
\newcommand{\leight}{DeepSeek-R1-Distill-Llama-8B\xspace}
\newcommand{\glmflash}{GLM-5.3-Flash\xspace}

\newcommand{\wfa}{W4A16\xspace}
\newcommand{\wta}{W3A16\xspace}
\newcommand{\wfae}{W4A8\xspace}
\newcommand{\nvfp}{NVFP4\xspace}
\newcommand{\wfaf}{W4A4\xspace}

\definecolor{avgbg}{gray}{0.93}
\mlsystitlerunning{QDA: Closed-Loop LLM Quantization Design with Agents}
\hypersetup{pdftitle={QDA: Closed-Loop LLM Quantization Design with Agents}, pdfauthor={Xingyang Li, Tianlang Zhao, Song Han, Ligeng Zhu}}

\begin{document}

\twocolumn[
\mlsystitle{QDA: Closed-Loop LLM \underline{Q}uantization \underline{D}esign with \underline{A}gents}

{\centering
\textbf{Xingyang Li}$^{*}$ \quad \textbf{Tianlang Zhao}$^{*}$ \quad \textbf{Song Han} \quad \textbf{Ligeng Zhu} \par\vskip 4pt
MIT \quad NVIDIA \par}

\vskip 0.2in
\begin{abstract}
Post-training quantization is the standard way to serve large language models at low cost.
However, the best quantization recipe differs across models and precisions.
As a result, experts must redesign it for every new model, which is time-consuming and often sub-optimal.
Fortunately, an LLM agent can take over this design by refining the recipe over rounds of building and evaluating candidates, so that more rounds lead to better recipes.
However, evaluating one quantized reasoning model takes tens of GPU-hours, and combining methods from different codebases means porting each to the new architecture by hand.
We present \textbf{\sys{}}, a system in which an agent designs the recipe for each model and precision.
To make evaluation fast, \sys{} scores each candidate in minutes by its output KL divergence from the original model on reasoning traces that the agent never sees, and checks promising ones with long generations to catch failures such as repetition loops.
To make building fast, \sys{} provides a library of public PTQ methods that are compatible with one another and with new model architectures.
Across four reasoning models, \sys{} improves accuracy over the strongest public recipe by \textbf{0.9}, \textbf{1.4} and \textbf{2.2} percentage points on average at \wfa{}, \wfae{} and \wta{}, using 64 GPU-hours per search.
On the 320B-parameter MoE model \glmflash{} at \nvfp{} W4A4, it recovers \textbf{98.3\%} of the FP8 model's accuracy, against \textbf{96.9\%} for the best vendor checkpoint.
Every \sys{} checkpoint also runs on existing serving kernels with no inference overhead.
We will release the library, the evaluator and the harness.

\end{abstract}
 ]

{\let\thefootnote\relax\footnotetext{$^{*}$Equal contribution.}}

\section{Introduction}
\label{sec:intro}
Large language models (LLMs) excel at mathematics, science and programming, and reasoning models go further by thinking for thousands of tokens before they answer~\cite{deepseekai2025r1,yang2025qwen3,chen2025longcot}.
However, serving them is costly in both memory and compute.
Post-training quantization (PTQ) is a standard way to cut both costs~\cite{frantar2023gptq,lin2024awq}.
For example, with its weights in \nvfp{}, the 320B-parameter \glmflash{}~\cite{zai2026glm53flash} fits on one GB300 GPU instead of the two that it needs in FP8.
Quantizing the activations as well lets matrix multiplications run on low-precision tensor cores, which are several times faster than BF16 ones~\cite{xiao2023smoothquant,lin2025qserve,nvidia2025nvfp4}.
\begin{figure}[t]
    \centering
    \includegraphics[width=\columnwidth]{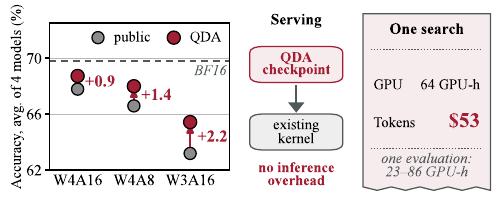}
    \vspace{-20pt}
    \caption{\textbf{\sys{} beats the strongest public recipe at low cost.} Left: average accuracy of four reasoning models at each precision, for the strongest public recipe and for \sys{}. Middle: \sys{} checkpoints run on existing kernels with no inference overhead. Right: the GPU-hours and agent tokens of one search.}
\label{fig:teaser}
\end{figure}

\begin{figure*}[t]
    \centering
    \includegraphics[width=\textwidth]{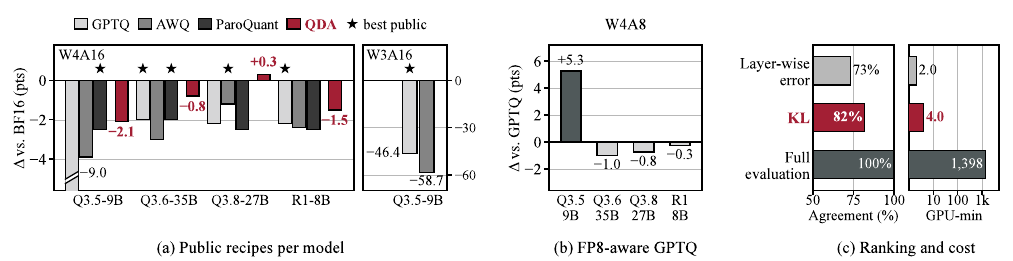}
    \vspace{-14pt}
    \caption{\textbf{The best recipe differs across models and precisions, and a full evaluation costs hundreds of times more than a KL score.}
(a)~Accuracy change from BF16 of three public recipes and \sys{} at \wfa{} on four models. A star marks the most accurate public recipe. On Q3.5-9B at \wta{}, AWQ and GPTQ swap order.
(b)~Accuracy change at \wfae{} of FP8-aware GPTQ, which fits GPTQ's codes to the kernel's FP8 scales instead of BF16 scales. This helps Q3.5-9B but hurts the other three.
(c)~Left: share of 44 checkpoint pairs that each measure ranks in the same order as accuracy. Right: GPU-minutes to compute each measure on one R1-8B checkpoint.
Q3.5-9B, Q3.6-35B, Q3.8-27B and R1-8B denote \qthreefive{}, \qthreesix{}, \qthreeeight{} and \leight{}.}
\label{fig:motivation}
\end{figure*}

Existing PTQ methods fix different sources of quantization error~\cite{xiao2023smoothquant,frantar2023gptq,lin2024awq}, so they can be stacked, and a combination often works better than any single method~\cite{ashkboos2024quarot,liu2025spinquant}.
A quantized checkpoint is therefore produced by a \emph{recipe}, which consists of the stacked methods, their settings and their calibration data (\refsec{sec:background:techniques}).
However, we find that the best recipe differs across models and precisions.
For example, with 4-bit weight-only quantization, the most accurate public recipe is ParoQuant~\cite{paroquant2026} on \qthreefive{}~\cite{qwen2026qwen35} but GPTQ~\cite{frantar2023gptq} on \leight{} (\Cref{fig:motivation}(a)).
On \qthreefive{} alone, AWQ~\cite{lin2024awq} beats GPTQ in 4-bit but loses to it in 3-bit weight-only quantization.

Each method is evaluated on only a few models and precisions, so the best recipe for a new model can only be found by trying candidates on it.
Experts therefore still decide by hand which methods to combine, with which settings and data.
For example, NVIDIA's ModelOpt keeps a separate hand-written recipe for each of 28 models, which records model-specific decisions such as which modules to leave in BF16~\cite{nvidia2024modelopt}.
Such hand design does not scale, because every new model needs its own recipes and new models arrive quickly.
Between October 2025 and September 2026, more than 90 open-weight models with over 10B parameters were released~\cite{epoch2026aimodels}, which would need over 360 recipes at our four precisions.

LLM coding agents already write code to improve algorithms, GPU kernels and computer systems~\cite{novikov2025alphaevolve,cao2026ksearch,adrs2025}.
Recipe design is a natural task for such an agent, because it also proceeds by trial and error, building and evaluating one candidate after another as an expert would.
However, two problems stand in the way.
First, evaluating a candidate is expensive.
Because a reasoning model writes thousands of tokens per answer and each problem must be sampled many times, a full evaluation of one quantized checkpoint takes up to 86 GPU-hours (\refsec{sec:motivation:measure}).
At this cost, a search can try only a few candidates, so a wider search needs a much cheaper proxy metric for accuracy.
Second, building a candidate is costly.
Most PTQ methods are released as separate research code, so a recipe that mixes them must be ported by hand to each new architecture, such as linear-attention hybrids~\cite{yang2025gateddeltanet,kimi2025linear} and fine-grained mixtures of experts~\cite{deepseekai2024v3,openai2025gptoss}.
Even a method that a toolkit supports can be too slow for a search (\refsec{sec:motivation:challenges}).
An agent that fills these gaps itself writes about 3k lines of quantization code per search and still ends with a less accurate checkpoint (\refsec{sec:eval:ablation}).

To this end, we present \textbf{\sys{}}, a system in which an agent designs the quantization recipe for each model and precision (\Cref{fig:overview}).
To address the first problem, we use the KL divergence of the candidate's next-token distributions from the original model's as the proxy metric, measured on \emph{reference traces}, reasoning traces that the original model wrote.
Because the original model's distributions are cached, scoring a candidate takes minutes, which is 350--990$\times$ cheaper than a full evaluation.
The agent also generates full answers for a few promising candidates to reveal rare failures such as repetition loops.
To address the second problem, we implement 21 PTQ methods in one library behind a common interface.
Every method takes a quantized checkpoint as input and returns one, so the agent can stack methods in any order.
Finally, a harness runs the search without human supervision and keeps the test benchmarks and the reference traces out of the agent's reach, because a recipe tuned on them would score higher without being any better~\cite{kapoor2025agents,xu2024contamination}.

On four reasoning models that span dense Transformers, linear-attention hybrids and mixture-of-experts (MoE) models, \sys{} improves the average accuracy over the strongest public recipe by \textbf{0.9}, \textbf{1.4} and \textbf{2.2} points at \wfa{}, \wfae{} and \wta{} (\Cref{fig:teaser}).
Each search uses 64 GPU-hours.
On the MoE model, \sys{} also outperforms greedy search and ASHA~\cite{li2020asha} by up to \textbf{1.2} points at \wfae{} and \wta{} when they search the same library with the same budget, and the same agent given ModelOpt and its recipe-search skill falls 0.5--1.3 points behind even when it selects on the test benchmarks.
On the 320B-parameter \glmflash{} at \nvfp{}, a single 6.5-hour search recovers \textbf{98.3\%} of the FP8 model's accuracy, against 96.9\% and 96.3\% for the checkpoints released by NVIDIA and Red Hat AI~\cite{nvidia2026glmnvfp4,redhat2026glmnvfp4}.
Moreover, every \sys{} checkpoint runs on the existing serving kernel of its format and adds \textbf{no inference overhead}.

\section{Background}
\label{sec:background}

\begin{table}[t]
  \centering
  \caption{Precisions evaluated. The first three run on GH200 with vLLM~0.27.1 for the four main models, and \nvfp{} runs on GB300 with vLLM~0.31 for \glmflash{}.}
\label{tab:precisions}
  \vskip 0.04in
  \footnotesize
  \resizebox{\columnwidth}{!}{
  \begin{tabular}{llll}
    \toprule
    \textbf{Precision} & \textbf{Weights} & \textbf{Activations} & \textbf{Kernel} \\
    \midrule
    \wfa{}  & INT4 g128 asym & BF16 & Marlin \\
    \wta{}  & \makecell[l]{INT3 g128 sym;\\ head + embeddings INT4 g64} & BF16 & Humming \\
    \wfae{} & INT4 g128 sym  & FP8 (E4M3) per-token & CUTLASS W4A8 \\
    \midrule
    \nvfp{} \wfaf{} & FP4 (E2M1) g16, FP8 scales & FP4 (E2M1) g16 & NVFP4 GEMM \\
    \bottomrule
  \end{tabular}
  }
\end{table}

\subsection{Quantized LLM Serving}
\label{sec:background:formats}
Quantization lowers both the memory and the compute of serving.
Smaller weights let a model fit on fewer GPUs and speed up decoding, which is bound by weight reads~\cite{lin2025qserve,lin2024awq}.
Quantized activations further let matrix multiplications use faster tensor cores, which speeds up compute-bound prefill and large-batch decoding.

\MyPara{Quantization.}
Quantization maps a group $\mathbf{x}$ of values to $b$-bit codes with a scale $s$ and a zero point $z$:
\begin{equation}
Q(\mathbf{x}) = s\,\Big(\mathrm{clamp}\big(\big\lfloor \tfrac{\mathbf{x}}{s} \big\rceil + z,\ 0,\ 2^b-1\big) - z\Big).
\label{eq:quant}
\end{equation}
Symmetric formats use signed codes without $z$, and \nvfp{}~\cite{nvidia2025nvfp4} rounds to 4-bit floating-point values instead of integers.
We write W$x$A$y$ for $x$-bit weights and $y$-bit activations, and \Cref{tab:precisions} lists our four formats.

\MyPara{Quantized matrix multiplication.}
A linear layer computes $\mathbf{Y}=\mathbf{X}\mathbf{W}^{\top}$, and its quantized kernel computes
\begin{equation}
\mathbf{Y} \approx Q(\mathbf{X})\,Q(\mathbf{W})^{\top},
\label{eq:qgemm}
\end{equation}
where $Q(\mathbf{W})$ is stored in the checkpoint and $Q(\mathbf{X})$, if any, is computed on the fly.
The formats differ in which operands the tensor cores multiply.
At \wfa{} and \wta{}, the kernel dequantizes the weights to BF16 in registers, which saves memory traffic but not arithmetic~\cite{frantar2024marlin,lin2024awq}.
At \wfae{}, the activations are also quantized per token to FP8~\cite{micikevicius2022fp8}, the only type that vLLM's \wfae{} kernel takes.
At \nvfp{} \wfaf{}, both operands are 4-bit floats with one scale per 16 values, which Blackwell tensor cores multiply directly~\cite{nvidia2025nvfp4,rouhani2023microscaling}.
Each kernel also fixes the precision of the scales, such as FP8 at \wfae{}, and a quantizer must match it exactly.

\subsection{Post-Training Quantization Methods}
\label{sec:background:techniques}
Post-training quantization (PTQ) quantizes a trained model without retraining it, using only a small set of calibration texts.
Most methods work one linear layer at a time and minimize its output error on the calibration inputs $\mathbf{X}$:
\begin{equation}
\min_{Q(\mathbf{W})} \; \big\lVert \mathbf{X}\mathbf{W}^{\top} - \mathbf{X}\,Q(\mathbf{W})^{\top} \big\rVert_F^2 .
\label{eq:layerobj}
\end{equation}
A recipe makes four choices: a transformation, a rounding method, the scales and their tuning, and the calibration data.

\MyPara{Transformations.}
Since $\mathbf{X}\mathbf{W}^{\top}=(\mathbf{X}\mathbf{T}^{-1})(\mathbf{W}\mathbf{T}^{\top})^{\top}$ for any invertible $\mathbf{T}$, a transformation can reshape both operands without changing the output, for example to tame the outliers in a few activation channels~\cite{xiao2023smoothquant}.
A diagonal $\mathbf{T}$ rescales channels: SmoothQuant moves outliers from the activations into the weights~\cite{xiao2023smoothquant}, and AWQ scales up the weight channels that meet large activations, protecting them from rounding~\cite{lin2024awq}.
An orthogonal $\mathbf{T}$, such as a Hadamard rotation, mixes all channels so that none stands out~\cite{ashkboos2024quarot,liu2025spinquant,xi2023int4}. ParoQuant combines learned pairwise rotations with channel scaling~\cite{paroquant2026}.

\MyPara{Rounding.}
Rounding each weight to the nearest code ignores how the errors of different weights add up in the output.
GPTQ~\cite{frantar2023gptq} solves \Cref{eq:layerobj} approximately: it rounds one column at a time and updates the columns not yet rounded to cancel the error, using the second-order statistics $\mathbf{X}^{\top}\mathbf{X}$ of the calibration inputs.
Learned rounding, such as AutoRound~\cite{cheng2024autoround} and GSQ~\cite{dadgarnia2026gsq}, instead learns the rounding of each weight by gradient descent, at a higher cost.

\MyPara{Scales and tuning.}
A larger scale covers the extreme values but rounds the others coarsely, so clipping searches for the scale that best balances the two.
Since \Cref{eq:layerobj} cannot see errors that build up across layers, tuning methods train the scales so that the quantized model matches the original, block by block~\cite{shao2024omniquant,chen2025efficientqat} or by distillation~\cite{malinovskii2024pvtuning}.

\MyPara{Calibration data.}
Every statistic above comes from the calibration inputs, so their choice alone changes the accuracy of the result~\cite{williams2024calibration}.
Generic text is the default, and text that the original model generates (self-generated data) can work better~\cite{williams2025selfcalib}.

\MyPara{Combining the steps.}
Because each choice removes a different error, applying them in sequence, for example a rotation followed by GPTQ, is often more accurate than any single method~\cite{ashkboos2024quarot,liu2025spinquant}.
Yet steps that treat the same error overlap, so the best combination cannot be predicted from each method alone.

\section{Motivation}
\label{sec:motivation}

\subsection{The Best Recipe Differs per Model and Precision}
\label{sec:motivation:transfer}
At \wfa{}, no public recipe is the most accurate on every model (\Cref{fig:motivation}(a)).
ParoQuant is the best on \qthreefive{}, AWQ on \qthreeeight{} and GPTQ on \leight{}.
On a single model, the ranking also changes with the precision.
On \qthreefive{}, AWQ is 5.1 points more accurate than GPTQ at \wfa{} but 12.3 points less accurate at \wta{}.
Even a single technique can help one model and hurt the others.
For example, the \wfae{} kernel stores each scale in FP8, whereas GPTQ computes with BF16 scales.
Running GPTQ with the FP8 scales instead (FP8-aware GPTQ) raises the accuracy of \qthreefive{} by 5.3 points but lowers it on the other three models (\Cref{fig:motivation}(b)).
A recipe found for one model and precision therefore does not carry over, and each new pair needs its own search.

\subsection{Evaluating a Recipe Is Expensive}
\label{sec:motivation:measure}
Earlier searches over quantization recipes could evaluate each candidate in minutes.
AutoML for convolutional networks measured accuracy on a small validation set~\cite{he2018amc,wang2019haq}, and previous LLM quantization methods were judged by short benchmarks that need only a few generated tokens per question~\cite{frantar2023gptq,lin2024awq}.

Reasoning models break this assumption.
They are graded on the final answer after a long chain of thought, which often runs to tens of thousands of tokens, and one different token can change the rest of the reasoning and the answer~\cite{yuan2025nondeterminism}.
A reliable estimate therefore averages many samples per problem, especially since AIME holds only tens of problems~\cite{hochlehnert2025sober}, so a full evaluation of one quantized checkpoint generates about ten thousand long answers and takes 23 GPU-hours on \leight{} and 60--86 GPU-hours on our Qwen models (\Cref{fig:motivation}(c)).
Fewer problems or samples make the estimate too noisy, and scoring every candidate on the test sets would overfit them~\cite{blum2015ladder}.
A search therefore needs a proxy metric that takes minutes yet reflects the whole served model.

\subsection{Building a Candidate Is Costly}
\label{sec:motivation:challenges}
Building a candidate is costly in two ways.
First, a recipe can use only methods whose code runs on the model's architecture.
Most PTQ methods are released as research code for the model families in their papers.
Toolkits such as llm-compressor, AutoRound and ModelOpt run only the methods that they have reimplemented and extend per-architecture tables by hand for each new model~\cite{llmcompressor2024,cheng2024autoround,nvidia2024modelopt}, so llm-compressor took three weeks after the release of Qwen3-Next to support calibration on it~\cite{llmcompressor2025qwen3next}.
Second, a build can be too slow for a search.
For example, one GPTQ build of the 256-expert \qthreesix{} with llm-compressor~0.13 takes about 16 GPU-hours, a quarter of the 64 GPU-hours that we give each search.
Without our library, the agent fills these gaps itself, writing about 3k lines of quantization code per search, and still ends with a less accurate checkpoint (\refsec{sec:eval:ablation}).
Our library fills them once for all searches.
It derives which layers feed which from the model itself, so a new architecture such as \glmflash{} needs no per-method changes, and it builds this GPTQ in 0.42 GPU-hours (\refsec{sec:system:lib}).

\section{\sys{}: Recipe Design by an LLM Agent}
\label{sec:design}
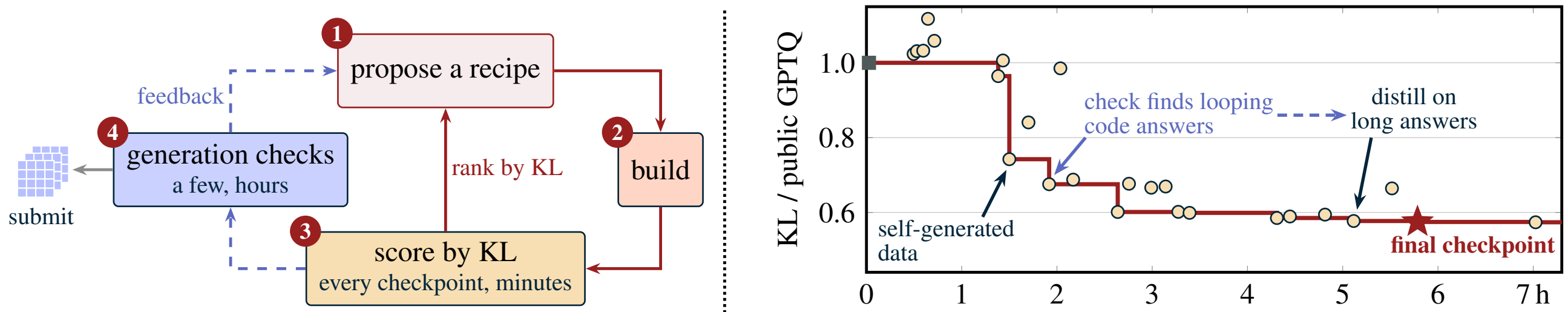
\begin{figure*}[!t]
\providecommand{\annot}{\fontsize{7.5}{8.5}\selectfont}
\centering
\begin{tikzpicture}[hfig,
  stp/.style={dbox, font=\footnotesize, minimum height=0.8cm, inner xsep=4pt},
  stepn/.style={circle, fill=hCard, text=white, font=\scriptsize\bfseries, inner sep=0.5pt, minimum size=0.34cm},
  fast/.style={->, draw=hCard, line width=0.9pt, >={Stealth[length=4.5pt]}},
  slow/.style={->, draw=cChkD, line width=0.9pt, dashed, >={Stealth[length=4.5pt]}},
  alab/.style={font=\annot, inner sep=1.5pt},
  scap/.style={font=\footnotesize, anchor=base, inner sep=0pt},
]
\colorlet{cChk}{hkBlue!40}\colorlet{cChkD}{hkBlue!75!black}
\newcommand{\wgrid}[2]{%
  \foreach \i in {0,...,3} \foreach \j in {0,...,3} {
    \fill[dBlue] ($#1+(\i*#2,\j*#2)$) rectangle ++(#2,#2);
    \draw[white, line width=0.4pt] ($#1+(\i*#2,\j*#2)$) rectangle ++(#2,#2);
  }}
\newcommand{\subl}{\annot\color{dLine}}
\def\yCap{0.1}\def\yCapB{-0.26}   %
\def\cx{4.8}\def\cy{2.0}\def\rx{2.35}\def\ry{1.08}
\node[stp, fill=hCard1, draw=hCard] (P) at (\cx,\cy+\ry) {propose a recipe};
\node[stp, fill=dOrange] (B) at (\cx+\rx,\cy) {build};
\node[stp, fill=dGreen] (K) at (\cx,\cy-\ry) {score by KL\\[-1pt]{\subl every checkpoint, minutes}};
\node[stp, fill=cChk] (C) at (\cx-\rx,\cy) {generation checks\\[-1pt]{\subl a few, hours}};
\foreach \n/\k in {P/1,B/2,K/3,C/4} \node[stepn] at (\n.north west) {\k};
\draw[fast] (P.east) -| (B.north);
\draw[fast] (B.south) |- (K.east);
\draw[fast] (K.north) -- node[alab, right, align=left, text=hCard] {rank by KL} (P.south);
\draw[slow] (K.west) -| (C.south);
\draw[slow] (C.north) |- node[alab, pos=0.3, left, text=cChkD] {feedback} (P.west);
\foreach \k in {2,1,0} {\wgrid{($(0.12,\cy-0.3)+(0.08*\k,0.08*\k)$)}{0.1}}
\draw[->, draw=hGrey, line width=0.9pt, >={Stealth[length=4.5pt]}] (C.west) -- (0.72,\cy);
\node[alab, anchor=north, align=center, text=dLine] at (0.38,\cy-0.35) {submit};
\node[scap] at (3.9,\yCap) {(a) The search loop};
\draw[black, line width=0.9pt, densely dotted] (7.85,-0.3) -- (7.85,3.75);
\begin{axis}[hplot, at={(9.4cm,0.88cm)}, anchor=south west, scale only axis,
  width=7.6cm, height=2.9cm, xmin=0, xmax=7.3, ymin=0.44, ymax=1.15,
  xtick={0,1,...,7}, xticklabels={0,1,2,3,4,5,6,7\,h}, ytick={0.6,0.8,1.0},
  yticklabels={0.6,0.8,1.0}, ylabel={KL / public GPTQ}, ylabel shift=-2pt,
  clip=false, axis line style={black, line width=0.9pt}]
\addplot[const plot, hCard, line width=1.3pt, mark=none] coordinates
  {(0.026,1.0) (1.383,0.9644) (1.500,0.7423) (1.918,0.6754) (2.638,0.6012) (3.392,0.5989) (4.309,0.5851) (5.115,0.5771) (5.785,0.5746) (7.025,0.5737) (7.3,0.5737)};
\addplot[only marks, mark=*, mark size=1.9pt, mark options={fill=dGreen, draw=dLine, line width=0.5pt}] coordinates
  {(0.495,1.0236) (0.529,1.0311) (0.596,1.0324) (0.646,1.1175) (0.713,1.0588) (1.383,0.9644) (1.433,1.0062) (1.500,0.7423) (1.701,0.8407) (1.918,0.6754) (2.036,0.9852) (2.170,0.6877) (2.638,0.6012) (2.756,0.6767) (2.990,0.6661) (3.141,0.6693) (3.274,0.6016) (3.392,0.5989) (4.309,0.5851) (4.446,0.5892) (4.814,0.5942) (5.115,0.5771) (5.517,0.6643) (7.025,0.5737)};
\addplot[only marks, mark=square*, mark size=2.0pt, mark options={fill=hSlate, draw=hSlate}] coordinates {(0.026,1.0)};
\node[font=\annot, anchor=south west, align=left, text=dLine, inner sep=1pt] (own) at (0.08,0.46) {self-generated\\[-1pt]data};
\draw[->, draw=dLine, line width=1pt, >={Stealth[length=5pt]}] (1.22,0.588) -- (1.47,0.714);
\node[font=\annot, anchor=south west, align=left, text=cChkD, fill=white, inner sep=1pt] (chk) at (2.25,0.80) {check finds looping\\[-1pt]code answers};
\draw[->, draw=cChkD, line width=1pt, >={Stealth[length=5pt]}] (2.3,0.80) -- (1.97,0.70);
\node[font=\annot, anchor=south, align=center, text=dLine, fill=white, inner sep=1pt] (dist) at (5.75,0.80) {distill on\\[-1pt]long answers};
\draw[->, draw=cChkD, line width=1pt, densely dashed, >={Stealth[length=5pt]}] (chk.east |- 0,0.86) -- (dist.west |- 0,0.86);
\draw[->, draw=dLine, line width=1pt, >={Stealth[length=5pt]}] (5.3,0.80) -- (5.15,0.607);
\coordinate (S) at (5.785,0.5746);
\node[font=\annot, anchor=north east, text=hCard, inner sep=1pt] at (7.27,0.548) {\textbf{final checkpoint}};
\end{axis}
\fill[hCard] ($(S)+(90:0.19)$) \foreach \k in {1,...,9} {-- ($(S)+({90+36*\k}:{mod(\k,2)==1 ? 0.08 : 0.19})$)} -- cycle;
\node[scap] at (12.9,\yCap) {(b) One search on Qwen3.6-35B-A3B at W4A16};
\end{tikzpicture}
\vspace{-10pt}
\caption{\textbf{The \sys{} loop.} Every candidate gets a KL score within minutes, a few promising ones also get a long-generation check, and the agent combines both to decide what to change next.
(a)~The four steps of one round. The numbers match steps (1)--(4) in \refsec{sec:design}.
(b)~The KL of every checkpoint in one search, relative to the public GPTQ checkpoint. The line shows the lowest KL so far. The blue label marks a long-generation check, and the black labels mark two changes that the agent made (\refsec{sec:design:doubt}).}
\label{fig:overview}
\end{figure*}

As shown in \Cref{fig:overview}(a), the agent designs a recipe in a loop of four steps.
In each round, (1)~the agent proposes a recipe, (2)~the library builds it into a candidate checkpoint, and (3)~the evaluator scores the checkpoint by its KL divergence from the original model, which takes minutes.
For a few promising candidates, (4)~the agent also generates long answers to find failures such as repetition loops.
When the budget runs out, the agent submits the checkpoint it considers best, and only this checkpoint is evaluated on the test benchmarks.
\refsec{sec:system} describes the system.

\subsection{Problem Setup}
\label{sec:design:setup}
A \emph{recipe} is an ordered list of methods from \refsec{sec:background:techniques}, each with its settings and calibration data, that turns the original model~$M$ into a quantized checkpoint.

\MyPara{Objective.}
The goal is the checkpoint that best preserves the general abilities of the original model.
We measure them by the mean accuracy $A$ of a full evaluation over math, science, code and instruction following, the categories that model releases report~\cite{deepseekai2025r1,yang2025qwen3}, and average the categories as recent quantization work does~\cite{paroquant2026}.
Within a GPU budget~$B$, the search looks for
\begin{equation}
\max_{i} \; A\big(x^{(i)}\big) \quad \text{s.t.} \quad \textstyle\sum_{i} c^{(i)} \le B,
\label{eq:problem}
\end{equation}
where $x^{(i)}$ is the $i$-th candidate and $c^{(i)}$ the GPU time for its data, build, score and checks.
Measuring $A$ for one checkpoint takes 23--86 GPU-hours on our models (\refsec{sec:motivation:measure}), so a search that measured $A$ for every candidate could afford only a few of them.
The search therefore ranks candidates by KL (step~3) and generation checks (step~4) and measures $A$ only on the final checkpoint.

\MyPara{Admissible recipes.}
A checkpoint must keep the public baselines' quantized modules, bit-width and format (\Cref{tab:precisions}) and run on their kernels with no new operator.
A public checkpoint may be refined but not submitted as is.

\MyPara{The agent.}
In step~(1), the agent writes a recipe.
It works in a sandbox with the library, a card per method that says what it does and when it fails, and a task statement (App.~\ref{app:prompts}), and it can write its own code, for example to generate calibration data with the original model.
Between proposals, it reads the log of its scores, the results of its checks and its own notes, and from them decides which step or setting to change next.

\subsection{Scoring Candidates by Their KL}
\label{sec:design:kl}
In step~(3), the evaluator scores a checkpoint by how much its next-token predictions differ from the original model's.

\MyPara{Why the KL.}
We start from the signal that most PTQ methods already compute, the layer-wise error of \Cref{eq:layerobj}, which GPTQ and AWQ minimize.
However, it looks at one layer at a time and cannot see the errors that build up across layers.
For example, on \qthreefive{} at \wfa{}, the layer-wise error prefers GPTQ over AWQ, although AWQ's $A$ is 5.1 points higher.
We therefore compare the whole quantized model with the original one, by the KL divergence between their next-token distributions.
The evaluator computes these distributions on the quantized checkpoint as vLLM serves it, so the KL also includes errors that only the serving kernel introduces, such as the FP8 scales of the \wfae{} kernel (\refsec{sec:motivation:transfer}).
To verify this choice, we take 44 pairs of public baselines and simple recipes at three precisions (\Cref{fig:motivation}(c)).
The KL orders 82\% of the pairs as their accuracy does, against 73\% for the layer-wise error.

\MyPara{The score.}
The evaluator holds a fixed set of \emph{reference traces}, which the original model wrote for prompts in math, science, code, and writing and planning that do not overlap with the test benchmarks.
It feeds each trace to the checkpoint and averages the per-token KL divergence over all generated positions~$\mathcal{T}$:
\begin{equation}
\mathrm{KL}(x) = \frac{1}{|\mathcal{T}|} \sum_{(s,t) \in \mathcal{T}} D_{\mathrm{KL}}\big(p_{M}(\cdot | s_{<t}) \,\|\, p_{x}(\cdot | s_{<t})\big),
\label{eq:kl}
\end{equation}
where $s_{<t}$ is the prefix of trace $s$ before position $t$.
Averaging within one domain gives the per-domain KL.

\MyPara{Estimating the KL from top tokens.}
Storing the original model's full distribution at every position would take 0.21--0.56\,TB per set of reference traces in BF16, since a vocabulary of 128k--248k tokens meets 0.49--1.13 million positions.
The evaluator therefore keeps, at each position $t$, only the $k=128$ most likely tokens $\mathcal{K}_t$ of the original model and one bucket for the rest, so each position contributes
\begin{equation}
\sum_{v \in \mathcal{K}_t} p_M(v)\log\frac{p_M(v)}{p_x(v)} \;+\; \bar p_M \log\frac{\bar p_M}{\bar p_x},
\label{eq:kltopk}
\end{equation}
where $\bar p = 1-\sum_{v\in\mathcal{K}_t} p(v)$.
\begin{proposition}\label{prop:topk}
The estimate of \Cref{eq:kltopk} never exceeds the full KL of \Cref{eq:kl}, and the two are equal when $p_M(v)/p_x(v)$ is the same for every token $v \notin \mathcal{K}_t$.
\end{proposition}
The estimate is the KL between the two distributions after all tokens outside $\mathcal{K}_t$ are merged into one bucket, and merging tokens can only lower the KL by the log-sum inequality~\cite{cover2006elements}.
Sparse logit distillation calls the same bucket a ghost token~\cite{anshumann2025sparselogit}.
In practice the bucket holds on average 0.015--0.048\% of the original model's probability, and on nine checkpoints that we scored both ways, the estimate is 0--0.4\% below the full KL.
The original model's top-$k$ distributions on the reference traces are computed once and cached, so scoring a candidate runs only its quantized checkpoint and takes minutes (\refsec{sec:system:judge}).

\MyPara{Marking ties.}
The evaluator marks two checkpoints as tied in the agent's log when their KLs differ by less than $\sigma$, so that the agent does not act on chance differences.
Here $\sigma$ is the relative standard deviation of the KL when one recipe is rebuilt on another random subset of its calibration data, 0.5--1.6\% on our models (App.~\ref{app:calibration}).
Scoring the same checkpoint again returns exactly the same value, so this variation comes from the build and not from the evaluator.

\begin{figure*}[t]
\centering
\providecommand{\annot}{\fontsize{7.5}{8.5}\selectfont}
\begin{tikzpicture}[hfig,
  panel/.style={draw=black, densely dotted, line width=0.9pt, rounded corners=3pt},
  banner/.style={fill=hCard, text=white, font=\footnotesize\bfseries, rounded corners=3pt, inner xsep=4.5pt, inner ysep=2pt, text height=1.6ex, text depth=0.45ex},
  big/.style={draw=dLine, line width=0.6pt, rounded corners=3pt, align=center, font=\footnotesize, inner sep=2pt},
  mstep/.style={big, fill=hCard1, draw=hCard, minimum height=0.8cm, minimum width=1.34cm},
  tile/.style={draw=dLine, line width=0.6pt, fill=hTeal1, font=\footnotesize, minimum width=0.95cm, minimum height=0.42cm, inner sep=0pt, text height=1.6ex, text depth=0.45ex},
  tool/.style={draw=dLine, line width=0.6pt, rounded corners=1.5pt, fill=white, font=\footnotesize\ttfamily, minimum width=1.3cm, minimum height=0.42cm, inner sep=0pt, text height=1.6ex, text depth=0.45ex},
  note/.style={font=\annot, text=dLine, inner sep=1pt, align=center},
  llab/.style={font=\footnotesize, text=hCard, inner sep=1.5pt},
  arr/.style={->, draw=dLine, line width=0.7pt, >={Stealth[length=4pt]}},
  carr/.style={->, draw=hCard, line width=0.8pt, >={Stealth[length=4.5pt]}},
  harr/.style={->, draw=hSlate, line width=0.8pt, dashed, >={Stealth[length=4.5pt]}},
]
\newcommand{\cells}[3]{\foreach \i in {1,...,#3} \foreach \j in {1,...,#3} {
    \pgfmathtruncatemacro{\c}{mod(\i*7+\j*3,4)}
    \ifnum\c=0 \def\cc{hTeal2}\else\def\cc{hTeal}\fi
    \fill[\cc] ($#1+(\i*#2-#2,\j*#2-#2)$) rectangle ++(#2,#2);
    \draw[white, line width=0.4pt] ($#1+(\i*#2-#2,\j*#2-#2)$) rectangle ++(#2,#2);}}
\newcommand{\alock}[1]{%
  \draw[dLine, line width=0.7pt] ($#1+(-0.06,0.045)$) arc[start angle=180, end angle=0, radius=0.06];
  \fill[dLine, rounded corners=0.5pt] ($#1+(-0.11,-0.085)$) rectangle ($#1+(0.11,0.055)$);}
\newcommand{\sref}[1]{{\normalfont\scriptsize\color{white}\S\ref*{#1}}}
\def\pad{0.24}\def\gap{1.6}
\def\rA{4.33}\def\rR{3.50}\def\rC{2.44}
\def\yB{1.83}\def\yP{4.89}
\def\yS{0.98}   %
\node[mstep, anchor=west] (s1) at (0.35,\rR) {\strut Transform};
\node[mstep, anchor=west] (s2) at ($(s1.east)+(0.22,0)$) {\strut Rounding};
\node[mstep, anchor=west] (s3) at ($(s2.east)+(0.22,0)$) {Scales \&\\[-1pt]tuning};
\draw[arr] (s1) -- (s2); \draw[arr] (s2) -- (s3);
\coordinate (ck) at ($(s3.east)+(0.27,-0.3)$);
\cells{(ck)}{0.15}{4}
\node[inner sep=0pt, fit={(ck)($(ck)+(0.6,0.6)$)}] (CK) {};
\draw[arr] (s3.east) -- (CK.west);
\node[note, anchor=south] (ckl) at ($(CK.north)+(0,0.05)$) {served\\[-2pt]checkpoint};
\node[tile] (a2) at (s2 |- {(0,\rA)}) {hybrid};
\node[tile, anchor=east] (a1) at (a2.west) {dense};
\node[tile, anchor=west] (a3) at (a2.east) {MoE};
\node[note, anchor=base east] (adl) at ($(a1.base west)+(-0.1,0)$) {adapters};
\draw[arr] (a2.south) -- (s2.north);
\node[big, fill=white, minimum height=0.74cm, anchor=west] (ke) at (s1.west |- {(0,\rC)}) {\strut$s \to \mathrm{fp8}(s)\cdot q$\\[-1pt]{\annot\strut FP8-aware GPTQ}};
\node[big, fill=white, minimum height=0.74cm, anchor=east, align=left, inner xsep=4pt] (be) at (ckl.east |- {(0,\rC)}) {\hspace{0.75cm}\strut batched GPTQ\\[-1pt]\hspace{0.75cm}{\annot\strut all experts at once}};
\foreach \k in {2,1,0}
  \fill[hTeal2, draw=dLine, line width=0.4pt] ($(be.west)+(0.14+0.08*\k,-0.22+0.08*\k)$) rectangle ++(0.3,0.3);
\node[inner sep=0pt, fit=(s1)(s3)(CK)(ckl)(a1)(a3)(adl)(ke)(be)] (cL) {};
\coordinate (Lw) at ($(cL.west)+(-\pad,0)$);  \coordinate (Le) at ($(cL.east)+(\pad,0)$);
\draw[panel] (Lw |- {(0,\yB)}) rectangle (Le |- {(0,\yP)});
\node[inner sep=0pt, fit={(Lw |- {(0,\yB)}) (Le |- {(0,\yP)})}] (PL) {};
\node[banner] at (PL.north) {Composable Library~~\sref{sec:system:lib}};
\coordinate (Sw) at ($(Le)+(\gap,0)$);
\node[tool, anchor=west] (tq) at ($(Sw |- {(0,\rR)})+(\pad,0)$) {\syscmd};
\node[tool, anchor=west] (tk) at ($(tq.east)+(0.16,0)$) {kl};
\node[tool] (ts) at (tq |- {(0,\yB+0.58)}) {qsched};
\node[tool] (th) at (tk |- ts) {handin};
\coordinate (am) at ($(tq.east |- {(0,\rA)})!0.5!(tk.west |- {(0,\rA)})$);
\node[big, fill=hCard1, draw=hCard, minimum height=0.58cm, minimum width=2.32cm] (ag) at ($(am)+(0,-0.14)$) {Coding agent};
\coordinate (hub) at ($(ag.south)!0.5!(ag.south |- tq.north)$);
\draw[dLine, line width=0.6pt] (ag.south) -- (hub);
\draw[dLine, line width=0.6pt] (tq.north |- hub) -- (tk.north |- hub);
\draw[dLine, line width=0.6pt] (tq.north |- hub) -- (tq.north);
\draw[dLine, line width=0.6pt] (tk.north |- hub) -- (tk.north);
\draw[dLine, line width=0.6pt] (tq.south) -- (tq.south |- ts.north);
\draw[dLine, line width=0.6pt] (tk.south) -- (tk.south |- th.north);
\node[note, fill=white, inner sep=1.5pt] (sbx) at ($(tq.south)!0.5!(th.north)$) {no GPU, no internet};
\node[inner sep=0pt, fit=(ag)(tq)(tk)(ts)(th)(sbx)] (cS) {};
\coordinate (Se) at ($(cS.east)+(\pad,0)$);
\draw[panel] (Sw |- {(0,\yB)}) rectangle (Se |- {(0,\yP)});
\node[inner sep=0pt, fit={(Sw |- {(0,\yB)}) (Se |- {(0,\yP)})}] (PS) {};
\node[banner] at (PS.north) {Agent Sandbox};
\coordinate (Jw) at ($(Se)+(\gap,0)$);
\node[big, fill=hGreyL!60, minimum width=1.45cm, minimum height=0.8cm, anchor=west] (fz) at ($(Jw |- {(0,\rR)})+(\pad,0)$) {frozen copy\\[-1pt]+ hash};
\node[big, fill=hCard1, draw=hCard, minimum width=1.6cm, minimum height=0.8cm, anchor=west] (rd) at ($(fz.east)+(0.3,0)$) {KL inside\\[-1pt]vLLM};
\draw[arr] (fz) -- (rd);
\coordinate (jm) at ($(fz.west)!0.5!(rd.east)$);
\node[big, fill=hTeal1, minimum height=0.74cm, align=left, inner xsep=4pt] (an) at (jm |- {(0,\rC)}) {\hspace{0.3cm}reference traces\\[-1pt]\hspace{0.3cm}+ BF16 cache};
\alock{($(an.west)+(0.2,0)$)}
\draw[arr] (rd.south |- an.north) -- (rd.south);
\node[tile, minimum width=0.72cm, fill=hCard1, draw=hCard, anchor=east] (d2) at (jm |- {(0,\rA)}) {sci.};
\node[tile, minimum width=0.72cm, fill=hCard1, draw=hCard, anchor=east] (d1) at (d2.west) {math};
\node[tile, minimum width=0.72cm, fill=hCard1, draw=hCard, anchor=west] (d3) at (d2.east) {code};
\node[tile, minimum width=0.72cm, fill=hCard1, draw=hCard, anchor=west] (d4) at (d3.east) {plan};
\draw[arr] (rd.north) -- (rd.north |- d3.south);
\node[inner sep=0pt, fit=(fz)(rd)(an)(d1)(d4)] (cJ) {};
\coordinate (Je) at ($(cJ.east)+(\pad,0)$);
\draw[panel] (Jw |- {(0,\yB)}) rectangle (Je |- {(0,\yP)});
\node[inner sep=0pt, fit={(Jw |- {(0,\yB)}) (Je |- {(0,\yP)})}] (PJ) {};
\node[banner] at (PJ.north) {Evaluator~~\sref{sec:system:judge}};
\coordinate (mid) at ($(ts.east)!0.5!(th.west)$);
\node[banner, anchor=west] (sb) at ($(Lw |- {(0,\yS)})+(0.15,0)$) {Scheduler~~\sref{sec:system:sched}};
\node[font=\footnotesize, inner sep=1pt, anchor=west, text height=1.6ex, text depth=0.45ex] (rf) at ($(sb.east)+(0.25,0)$) {one queue for all jobs};
\node[font=\footnotesize, inner sep=1pt, anchor=west, text height=1.6ex, text depth=0.45ex] (gl) at ($(rf.east)+(0.35,0)$) {8 GPUs};
\foreach \g in {0,...,7} {\pgfmathsetmacro{\gx}{0.08+\g*0.26+0.1*int(\g/4)}
  \node[draw=hSlate, fill=hGreyL, line width=0.4pt, minimum size=0.22cm, inner sep=0pt, anchor=west] (g\g) at ($(gl.east)+(\gx,0)$) {};}
\draw[panel] (Lw |- {(0,\yS-0.27)}) rectangle ($(mid |- {(0,\yS+0.27)})+(-0.1,0)$);
\node[inner sep=0pt, fit={(Lw |- {(0,\yS-0.27)}) ($(mid |- {(0,\yS+0.27)})+(-0.1,0)$)}] (PH) {};
\node[big, fill=hGreyL!60, draw=hGrey, inner sep=0pt, fit={($(mid |- {(0,\yS-0.27)})+(0.1,0)$) (Je |- {(0,\yS+0.27)})}] (tb) {};
\node[font=\footnotesize, inner sep=0pt, text height=1.6ex, text depth=0.45ex] (tbt) at ($(tb)+(0.18,0)$) {Final evaluation on the test benchmarks};
\alock{($(tbt.west)+(-0.25,0)$)}
\draw[carr] ($(PS.west |- tq)+(0,0.09)$) -- ($(PL.east |- tq)+(0,0.09)$);
\draw[carr] ($(CK.east |- tq)+(0,-0.09)$) -- ($(tq.west)+(0,-0.09)$);
\node[llab, above] at ($(Le |- {(0,\rR+0.09)})+(\gap/2,0)$) {recipe};
\node[llab, below] at ($(Le |- {(0,\rR-0.09)})+(\gap/2,0)$) {checkpoint};
\draw[carr] ($(tk.east)+(0,0.09)$) -- ($(fz.west)+(0,0.09)$);
\draw[carr] ($(PJ.west |- tk)+(0,-0.09)$) -- ($(Se |- tk)+(0,-0.09)$);
\node[llab, above] at ($(Se |- {(0,\rR+0.09)})+(\gap/2,0)$) {checkpoint};
\node[llab, below] at ($(Se |- {(0,\rR-0.09)})+(\gap/2,0)$) {KL};
\draw[carr] (ts.south) -- (ts.south |- PH.north);
\coordinate (lb) at ($(ts.south |- PS.south)!0.5!(ts.south |- PH.north)+(0,-0.1)$);
\node[llab, anchor=base east] at ($(ts.south |- lb)+(-0.04,0)$) {jobs};
\draw[harr] (th.south) -- (th.south |- tb.north);
\node[llab, anchor=base west, text=hSlate] at ($(th.south |- lb)+(0.04,0)$) {submit};
\fitcheck{PL}{s1,s2,s3,CK,ckl,a1,a2,a3,adl,ke,be}
\fitcheck{PJ}{fz,rd,an,d1,d4}
\fitcheck{PS}{ag,tq,tk,ts,th}
\fitcheck{PH}{sb,rf,gl,g0,g7}
\end{tikzpicture}
\vspace{-6pt}
\caption{\textbf{System architecture of \sys{}.} \sys{} has three components: the composable library (\refsec{sec:system:lib}), the isolated evaluator (\refsec{sec:system:judge}) and the harness (\refsec{sec:system:iso}), which runs the agent in a sandbox and schedules its GPU jobs. From the sandbox, the agent can reach the rest of the system only through four commands: (1)~\texttt{\syscmd} builds a recipe in the library and returns its checkpoint, (2)~\texttt{kl} sends a checkpoint to the evaluator and returns its KL, (3)~\texttt{qsched} submits other jobs, such as long-generation checks, to the scheduler, and (4)~\texttt{handin} submits the final checkpoints, which are evaluated on the test benchmarks after the search.}
\label{fig:arch}
\end{figure*}
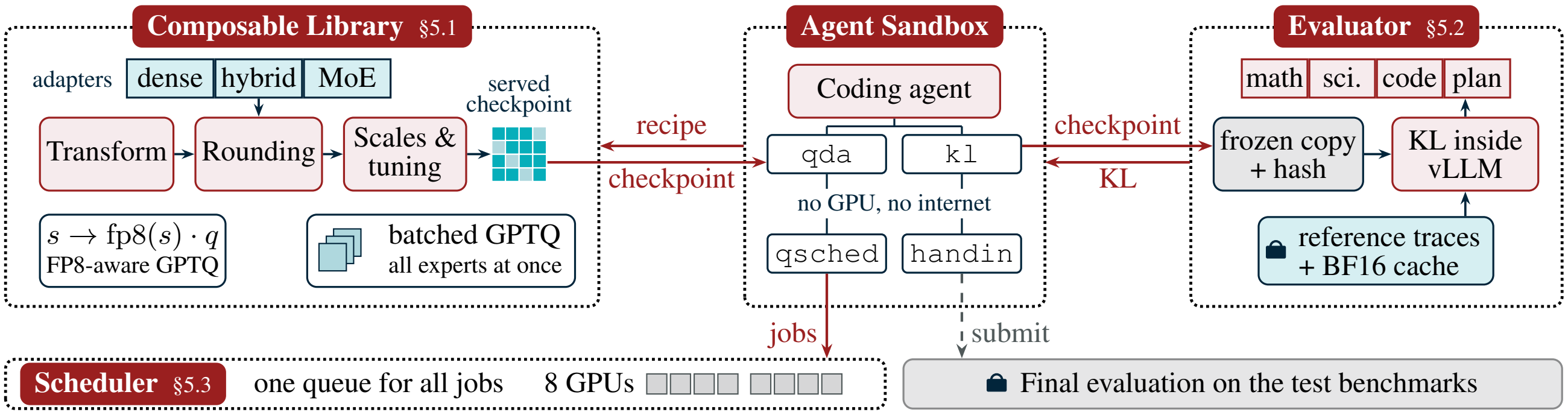

\subsection{Checking Long Answers}
\label{sec:design:doubt}
Only the answer after the thinking is graded, so a model that loops or never finishes its reasoning loses the whole problem.
Quantization can make this more likely, as quantized reasoning models can think longer than the original model~\cite{lotfi2026quantized}.
These failures tend to appear late in long answers, so they show up only when a checkpoint writes complete answers itself.
In step~(4), the agent therefore picks a candidate and has it write full answers to hard problems that the agent chooses itself, none of which come from the test benchmarks.
It counts the answers that are wrong, loop or are cut off, and weighs these counts together with the KL to decide what to change.

For example, on \qthreesix{} at \wfa{}, calibrating GPTQ on self-generated data lowered the KL by 30\% (\Cref{fig:overview}(b)).
In a long-generation check, however, this checkpoint cut off 12 of 191 answers at the length limit, against 3 for the original model.
Most of them were code answers that looped on debugging steps.
The agent noted that ``the truncation count is the signal I watch'' and, one minute after grading the check, began generating 600 hard prompts for long answers.
It kept the calibration on the original model's traces and added scale distillation on 1,200 of the quantized model's own long answers, and its final checkpoint beats the best public recipe by 1.3 points.

\section{The \sys{} System}
\label{sec:system}
\label{sec:impl}
\label{sec:system:agent}

A search must build and score dozens of checkpoints within its budget, safely and without human intervention.
\sys{} makes this possible with three components, a composable library, an evaluator inside the serving engine and a harness that runs the search (\Cref{fig:arch}).
The agent can write any code in its sandbox, but every GPU job, including its own code, goes through the harness, so that one scheduler shares the GPUs of a search and keeps to its budget (\refsec{sec:system:sched}).
The agent reaches the three components through four commands, \texttt{\syscmd} to build a recipe, \texttt{kl} to score a checkpoint, \texttt{qsched} for any other GPU job and \texttt{handin} to submit its final checkpoints.

\subsection{Composable Library}
\label{sec:system:lib}
\label{sec:system:exact}
Our library extends llm-compressor~\cite{llmcompressor2024} to 21 methods covering \refsec{sec:background:techniques} (\Cref{tab:library}).

\MyPara{Composing methods across architectures.}
Applying any method to a checkpoint that vLLM can load produces a new checkpoint that vLLM can also load.
Methods therefore stack as the steps of a recipe, and a method can also refine a public checkpoint or one that the agent built itself.
Methods such as AWQ and rotations must know which normalization layer feeds which linear layers, and research code hardcodes these module names for each model family.
Our library instead derives these relations from the model itself and checks each of them numerically, so a method written once runs on new architectures.
After the library was built, supporting \glmflash{}, a 288-expert model unseen by our main searches, took a 72-line adapter and an extension for its FP8 release and multi-token-prediction layer. Every method then ran on it, and the search needed no other development (\refsec{sec:eval:new}, App.~\ref{app:library}).

\MyPara{Matching the serving kernel.}
A kernel can store a quantity at lower precision than the quantizer assumes.
For example, vLLM's \wfae{} kernel casts each scale to FP8, whereas llm-compressor picks the codes for BF16 scales.
The library picks the codes for the scales that the kernel stores.

\MyPara{Fast builds.}
In llm-compressor~0.13, GPTQ solves each linear layer in turn, so one layer of 256 experts runs 768 small solves one after another.
The experts of a layer are independent problems of the same shape, so the library stacks their second-order statistics and solves all of them with one batched factorization.
One GPTQ build of \qthreesix{} thus takes 0.42 GPU-hours instead of about 16, with a KL within 1\% of llm-compressor's.

\subsection{Evaluator}
\label{sec:system:judge}
\Cref{eq:kl} needs both models' full-vocabulary next-token distributions at about a million positions.
A naive evaluator would run both models for every score.
Moreover, a serving engine such as vLLM returns only the few largest probabilities at each position.
The evaluator thus caches the original model's top-$k$ distributions once (about 1\,GB) and computes the candidate's side inside the engine.

\MyPara{Reducing inside the engine.}
The evaluator hooks vLLM's prompt log-probabilities inside its workers and reduces each full-vocabulary distribution to the terms of \Cref{eq:kltopk} in place, so only running sums leave the engine.
A score thus runs only the checkpoint, on the final evaluation's kernels, and is 350--990$\times$ cheaper than a full evaluation (App.~\ref{app:calibration}).

\subsection{Harness}
\label{sec:system:iso}
\label{sec:system:sched}
A search runs without human intervention, so the harness itself must keep the evaluation honest and the GPUs busy.

\MyPara{Isolating the evaluator.}
If the agent could read its evaluator, it could lower the KL without improving the model, for example by calibrating on the reference traces themselves~\cite{lin2024awq}.
The harness therefore keeps the reference traces and the test benchmarks out of the agent's reach.
The agent works in a sandbox with no GPU and no network except the model API, and a supervisor outside it checks and logs every tool call (App.~\ref{app:prompts}).
The evaluator runs outside the sandbox from a read-only copy of its code.
Each scoring job reads a frozen copy of the checkpoint and returns nothing but KL values.
These are the overall KL, the KL per domain and the KL on human-written text.
Every file the agent can read is scanned for test items beforehand, so that no test problem can reach the agent through calibration data or prompts (App.~\ref{app:calibration}).

\MyPara{Scheduling.}
A search runs dozens of GPU jobs, from KL scores of a few minutes to builds of several hours, and the agent submits them one by one as it works.
The agent waits for each KL score before its next step, but scores and builds run on the same GPUs.
Since the agent has no GPU in its sandbox, the harness holds these GPUs for the whole budget and runs every job from one priority queue.
A job states how many GPUs it needs, how long it expects to run and which jobs it waits for.
The harness copies the job's script when the job is submitted, so the agent can keep editing its files while the job waits or runs, and it runs the job in a container on GPUs that no other job uses.
KL scores go ahead of queued builds without stopping running ones, and no job starts if its expected duration runs past the deadline.
Over the 12 searches, a KL score starts a median of 5\,s after the agent asks, and jobs keep the GPUs busy 72\% of the time.

\section{Evaluation}
\label{sec:eval}
\begin{table*}[t]
\centering
  \caption{\textbf{Main results.} \sys{} improves the average accuracy over the strongest public recipe by 0.9, 1.4 and 2.2 points at \wfa{}, \wfae{} and \wta{}, and by up to 6.1 points (\leight{} at \wta{}).
Accuracy (\%) of four reasoning models on math, GPQA-Diamond (GPQA), LiveCodeBench (LCB) and IFBench (IFB; IFEval for R1-Llama), and their average (Avg).
Each \sys{} row is the final checkpoint of one 64-GPU-hour search, and the \qthreesix{} rows are means of three searches.
In each Avg column, the best quantized result is in bold and the second best is underlined.
Within each precision, all rows except ParoQuant, which adds an online rotation, run on the same kernel (\Cref{tab:precisions}).
QuaRot+GPTQ uses only QuaRot's offline rotations (R1 on the Qwen models, R1 and R2 on R1-Llama). At \wta{}, all Qwen searches but one on \qthreesix{} start from the GSQ checkpoint in this table and refine it.}
\label{tab:main}
  \vskip 0.08in
\resizebox{\textwidth}{!}{
\setlength{\tabcolsep}{2pt}
\footnotesize
\begin{tabular}{ll ccccc ccccc ccccc ccccc}
\toprule
\multirow{2}[3]{*}{\textbf{Prec.}} & \multirow{2}[3]{*}{\textbf{Method}} & \multicolumn{5}{c}{\qthreefive{}} & \multicolumn{5}{c}{\qthreesix{}} & \multicolumn{5}{c}{\qthreeeight{}} & \multicolumn{5}{c}{\leight{}} \\
\cmidrule(lr){3-7} \cmidrule(lr){8-12} \cmidrule(lr){13-17} \cmidrule(lr){18-22}
 & & \scriptsize{Math} & \scriptsize{GPQA} & \scriptsize{LCB} & \scriptsize{IFB} & \cellcolor{avgbg}\scriptsize{\textbf{Avg}} & \scriptsize{Math} & \scriptsize{GPQA} & \scriptsize{LCB} & \scriptsize{IFB} & \cellcolor{avgbg}\scriptsize{\textbf{Avg}} & \scriptsize{Math} & \scriptsize{GPQA} & \scriptsize{LCB} & \scriptsize{IFB} & \cellcolor{avgbg}\scriptsize{\textbf{Avg}} & \scriptsize{Math} & \scriptsize{GPQA} & \scriptsize{LCB} & \scriptsize{IFB} & \cellcolor{avgbg}\scriptsize{\textbf{Avg}} \\
\midrule
BF16 & --
  & 77.0 & 82.3 & 70.9 & 66.7 & \cellcolor{avgbg}74.2
  & 86.1 & 85.4 & 81.6 & 66.0 & \cellcolor{avgbg}79.8
  & 80.1 & 82.7 & 87.2 & 67.9 & \cellcolor{avgbg}79.5
  & 30.5 & 49.0 & 38.9 & 63.8 & \cellcolor{avgbg}45.6 \\
\midrule
\multirow{4}{*}{\wfa{}} & GPTQ
  & 62.5 & 78.6 & 53.2 & 66.7 & \cellcolor{avgbg}65.3
  & 82.8 & 83.9 & 79.8 & 64.5 & \cellcolor{avgbg}\underline{77.7}
  & 79.2 & 81.3 & 81.6 & 66.8 & \cellcolor{avgbg}77.2
  & 28.3 & 47.5 & 36.8 & 60.9 & \cellcolor{avgbg}\underline{43.3} \\
 & AWQ
  & 72.8 & 79.6 & 63.5 & 65.5 & \cellcolor{avgbg}70.4
  & 80.9 & 83.5 & 79.5 & 63.1 & \cellcolor{avgbg}76.8
  & 78.3 & 82.1 & 85.2 & 67.6 & \cellcolor{avgbg}\underline{78.3}
  & 28.1 & 48.1 & 36.6 & 59.8 & \cellcolor{avgbg}43.1 \\
 & ParoQuant
  & 73.9 & 80.2 & 68.7 & 64.2 & \cellcolor{avgbg}\underline{71.8}
  & 82.7 & 84.0 & 79.2 & 65.1 & \cellcolor{avgbg}\underline{77.7}
  & 75.6 & 80.9 & 83.5 & 67.6 & \cellcolor{avgbg}76.9
  & 28.3 & 46.7 & 34.9 & 62.4 & \cellcolor{avgbg}43.0 \\
 & \textbf{\sys{}}
  & 74.5 & 80.0 & 68.3 & 65.7 & \cellcolor{avgbg}\textbf{72.1}
  & 84.5 & 85.0 & 81.0 & 65.3 & \cellcolor{avgbg}\textbf{79.0}
  & 82.7 & 81.9 & 86.3 & 68.2 & \cellcolor{avgbg}\textbf{79.8}
  & 27.6 & 48.1 & 37.8 & 62.7 & \cellcolor{avgbg}\textbf{44.1} \\
\midrule
\multirow{4}{*}{\wfae{}} & GPTQ
  & 52.3 & 68.7 & 42.0 & 65.8 & \cellcolor{avgbg}57.2
  & 81.7 & 83.0 & 78.7 & 66.1 & \cellcolor{avgbg}\underline{77.4}
  & 76.7 & 82.0 & 83.6 & 67.6 & \cellcolor{avgbg}77.5
  & 28.0 & 46.9 & 36.3 & 60.8 & \cellcolor{avgbg}\underline{43.0} \\
 & AWQ
  & 57.0 & 68.9 & 36.2 & 62.8 & \cellcolor{avgbg}56.2
  & 77.8 & 81.0 & 74.2 & 64.7 & \cellcolor{avgbg}74.4
  & 73.9 & 80.6 & 82.8 & 67.4 & \cellcolor{avgbg}76.2
  & 25.7 & 45.2 & 33.9 & 59.2 & \cellcolor{avgbg}41.0 \\
 & QuaRot+GPTQ
  & 71.5 & 73.6 & 59.7 & 63.4 & \cellcolor{avgbg}\underline{67.1}
  & 81.4 & 82.4 & 79.5 & 64.6 & \cellcolor{avgbg}77.0
  & 80.6 & 80.6 & 86.2 & 68.1 & \cellcolor{avgbg}\underline{78.9}
  & 25.6 & 43.5 & 36.5 & 60.6 & \cellcolor{avgbg}41.6 \\
 & \textbf{\sys{}}
  & 74.6 & 81.3 & 68.5 & 65.1 & \cellcolor{avgbg}\textbf{72.3}
  & 82.2 & 83.6 & 79.3 & 65.0 & \cellcolor{avgbg}\textbf{77.5}
  & 81.8 & 80.3 & 86.0 & 68.3 & \cellcolor{avgbg}\textbf{79.1}
  & 27.5 & 46.7 & 36.5 & 61.5 & \cellcolor{avgbg}\textbf{43.1} \\
\midrule
\multirow{4}{*}{\wta{}} & GPTQ
  & 19.3 & 41.9 & 9.2 & 40.6 & \cellcolor{avgbg}27.8
  & 61.2 & 76.6 & 65.7 & 53.0 & \cellcolor{avgbg}64.1
  & 75.1 & 75.3 & 82.7 & 66.8 & \cellcolor{avgbg}75.0
  & 20.3 & 35.9 & 27.8 & 51.4 & \cellcolor{avgbg}\underline{33.9} \\
 & AWQ
  & 2.4 & 26.7 & 1.7 & 31.2 & \cellcolor{avgbg}15.5
  & 26.1 & 56.1 & 14.6 & 46.5 & \cellcolor{avgbg}35.8
  & 67.5 & 67.5 & 69.5 & 64.6 & \cellcolor{avgbg}67.3
  & 10.2 & 27.4 & 17.2 & 41.5 & \cellcolor{avgbg}24.1 \\
 & GSQ
  & 67.5 & 70.5 & 59.3 & 63.7 & \cellcolor{avgbg}\underline{65.2}
  & 80.5 & 82.4 & 76.4 & 64.5 & \cellcolor{avgbg}\textbf{75.9}
  & 81.9 & 78.6 & 83.6 & 66.7 & \cellcolor{avgbg}\underline{77.7}
  & 14.7 & 33.0 & 22.9 & 38.7 & \cellcolor{avgbg}27.3 \\
 & \textbf{\sys{}}
  & 70.7 & 73.5 & 63.5 & 59.3 & \cellcolor{avgbg}\textbf{66.8}
  & 81.8 & 82.2 & 77.0 & 62.0 & \cellcolor{avgbg}\underline{75.7}
  & 85.9 & 79.1 & 84.7 & 66.8 & \cellcolor{avgbg}\textbf{79.1}
  & 25.1 & 42.5 & 32.4 & 60.2 & \cellcolor{avgbg}\textbf{40.0} \\
\bottomrule
\end{tabular}
}
\end{table*}

\subsection{Setup}
\label{sec:eval:setup}

\MyPara{Models and precisions.}
We evaluate four reasoning models (\Cref{tab:models}), which we call Qwen3.5, Qwen3.6, Qwen3.8 and R1-Llama in figures and \Cref{tab:ablation}b.
Two of them are dense Gated DeltaNet hybrids (\qthreefive{}, \qthreeeight{}), one is a hybrid MoE with 256 experts (\qthreesix{}), and one is the dense \leight{}.
We quantize the same modules as the public baselines (App.~\ref{app:protocol}).
We use the formats and vLLM kernels of \Cref{tab:precisions} on GH200 GPUs, and we also run one search on \glmflash{} at \nvfp{} \wfaf{}, a pair that no main search saw (\refsec{sec:eval:new}).

\MyPara{Benchmarks.}
The test benchmark has four categories, and we report their mean.
They are math (AIME~2025 and HMMT~Feb~2026), science (GPQA-Diamond~\cite{rein2023gpqa}), code (LiveCodeBench~v6~\cite{jain2025livecodebench}) and instruction following (IFBench~\cite{pyatkin2025ifbench}), with 32, 16, 4 and 16 samples per problem (App.~\ref{app:protocol}).
\leight{} uses IFEval~\cite{zhou2023ifeval} instead of IFBench, since its accuracy of 20\% on IFBench is too low to separate recipes.
Models generate up to 81{,}920 tokens with the sampling settings of their model cards (App.~\ref{app:protocol}).
A sample that never ends its thinking scores 0.

\MyPara{Baselines.}
The public recipes are GPTQ~\cite{frantar2023gptq} and AWQ~\cite{lin2024awq} with the defaults of llm-compressor, QuaRot's offline rotations followed by default GPTQ (QuaRot+GPTQ)~\cite{ashkboos2024quarot} at \wfae{}, and two recent methods that report state-of-the-art accuracy, ParoQuant~\cite{paroquant2026} at \wfa{} and GSQ~\cite{dadgarnia2026gsq} at \wta{}.
GPTQ is asymmetric at \wfa{}. ParoQuant uses its released checkpoints. GSQ uses its released checkpoint on \qthreeeight{}, and on the other three models we run its official code ourselves (App.~\ref{app:baselines}).
Both ParoQuant and GSQ train the quantized model.
ParoQuant takes about 9 GPU-hours on an 8B model~\cite{paroquant2026}, and GSQ took 30--52 GPU-hours per model on our GPUs, close to the 64 GPU-hours of one search.

\MyPara{Agents and budget.}
The agent is Claude Code v2.1.303 with Claude Opus~5.5 at high effort, and one task statement serves every model and precision (App.~\ref{app:prompts}).
Each model--precision pair gets one search with 8 GH200 GPUs for 8 hours (64 GPU-hours) and \$19--75 of agent tokens, and the final evaluation with vLLM~0.27.1~\cite{kwon2023vllm} is not charged to it.
On \qthreesix{}, which the ablations also use, each precision gets three searches, and we report their mean (standard deviation 0.1--0.2 points).
Like a human expert, the agent may start from a public checkpoint (\refsec{sec:design:setup}).

\subsection{Accuracy Results}
\label{sec:eval:main}
\begin{table}[t]
  \centering
  \caption{\textbf{\glmflash{} at \nvfp{}.} \sys{} recovers 98.3\% of the FP8 accuracy on average, against 96.9\% and 96.3\% for the \nvfp{} releases of NVIDIA and Red Hat AI. Parentheses give the number of problems and samples per problem. Bold and underline mark the best and second-best \nvfp{} results.}
\label{tab:glm}
  \vskip 0.08in
  \footnotesize
  \setlength{\tabcolsep}{3pt}
  \begin{tabular}{lcccc}
    \toprule
    \textbf{Benchmark} & FP8 & NVIDIA & Red Hat AI & \textbf{\sys{}} \\
    \midrule
    Math (93, $\times$16) & 84.8 & \underline{80.9} & 78.8 & \textbf{83.3} \\
    GPQA-D (198, $\times$12) & 86.1 & 83.3 & \textbf{84.0} & \underline{83.9} \\
    LCB v6 (454, $\times$4) & 71.3 & \underline{69.3} & 69.2 & \textbf{70.4} \\
    SuperGPQA (9k, $\times$1) & 68.7 & \underline{67.8} & 67.4 & \textbf{67.9} \\
    \midrule
    Avg & 77.7 & \underline{75.3} & 74.8 & \textbf{76.4} \\
    \% of FP8 & 100 & \underline{96.9} & 96.3 & \textbf{98.3} \\
    \bottomrule
  \end{tabular}
\end{table}

As \Cref{tab:main} shows, \sys{} improves the average accuracy over the strongest public recipe of each pair by 0.9, 1.4 and 2.2 points at \wfa{}, \wfae{} and \wta{}.

\MyPara{Across models.}
\sys{} is the most accurate recipe on the dense model and the two Gated DeltaNet hybrids at every precision.
On \qthreeeight{} at \wfa{}, it is even 0.3 points above BF16, and on \leight{} at \wta{}, it is 6.1 points above GPTQ.
On the MoE model \qthreesix{}, \sys{} leads by 1.3 and 0.1 points at \wfa{} and \wfae{} and is 0.2 points below GSQ at \wta{}.

\MyPara{Across precisions.}
At \wfae{}, the gain is largest on \qthreefive{}, where default GPTQ and AWQ lose 17--18 points to BF16.
QuaRot+GPTQ recovers 9.9 of them with a Hadamard rotation, and FP8-aware GPTQ recovers 5.3 (\Cref{fig:motivation}(b)).
\sys{}'s recipe adds self-generated data, an FP8 scale search and scale distillation to the rotation, and it is 5.2 points more accurate than QuaRot+GPTQ.

\begin{table}[t]
  \centering
  \caption{\textbf{\sys{} beats the same agent with ModelOpt's workflow and tools} by 0.5--1.3 points on \qthreesix{}, with the same 64 GPU-hours. Four-category mean accuracy (\%).}
  \label{tab:modelopt}
  \vskip 0.04in
  \footnotesize
  \begin{tabular}{lccc}
    \toprule
    \textbf{Prec.} & Best public & Agent + ModelOpt & \textbf{\sys{}} \\
    \midrule
    \wfa{}  & 77.7 & \underline{78.5} & \textbf{79.0} \\
    \wfae{} & \underline{77.4} & 76.2 & \textbf{77.5} \\
    \wta{}  & \textbf{75.9} & 74.5 & \underline{75.7} \\
    \bottomrule
  \end{tabular}
\end{table}

\MyPara{A strong agent alone is not enough.}
We give the same agent NVIDIA's ModelOpt and its recipe-search skill~\cite{nvidia2026modeloptskill} instead of our library, evaluator and scheduler (App.~\ref{app:runtime}).
Although it selects candidates on the test benchmarks themselves, it is 0.5--1.3 points less accurate than \sys{} and even falls below the best public recipe at \wfae{} and \wta{} (\Cref{tab:modelopt}).
The gain of \sys{} comes from its search flow and infrastructure, not the agent alone.

\begin{figure*}[t]
    \centering
    \includegraphics[width=\textwidth]{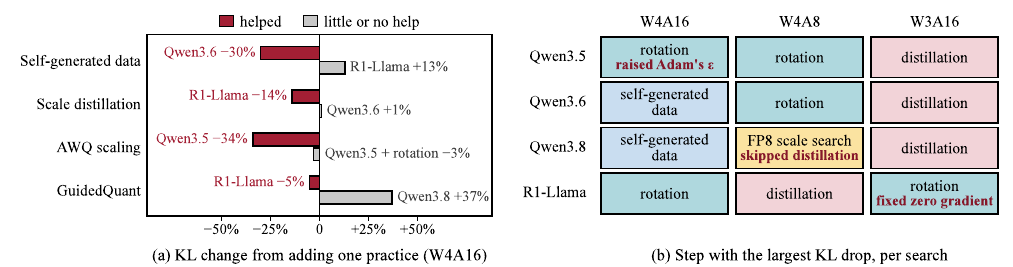}
    \vspace{-14pt}
    \caption{\textbf{What the searches found.} No published practice helps every model, and the step that lowers the KL most differs from search to search.
(a)~Change in KL at \wfa{} when one published practice is added to a recipe, measured between two checkpoints of the same search whose recipes differ only in that practice. Lower is better. Each practice lowers the KL on one model but not on another.
(b)~The step with the largest KL drop in each of the 12 final recipes. Bold notes mark problems that the agent diagnosed itself.}
\label{fig:search-findings}
\end{figure*}

\begin{table*}[t]
  \centering
  \caption{\textbf{Ablation and a fixed recipe.}
(a)~Replacing the agent with a predefined search algorithm (greedy search, ASHA) or replacing the library and scheduler with public tools and a plain GPU queue (w/o infra.), on \qthreesix{} with the same 64 GPU-hours, KL score and submission rule. \sys{} and w/o infra.\ are means of three searches.
(b)~A fixed recipe that came out of \sys{}'s searches: the pattern that five final recipes follow, applied with fixed settings and no search. It calibrates GPTQ on 640 answers that the original model wrote and then runs two rounds of scale distillation.
Four-category mean accuracy (\%). In each row, the best quantized result is in bold and the second best underlined.}
  \label{tab:ablation}
  \vskip 0.04in
  \footnotesize
  \setlength{\tabcolsep}{3.5pt}
  \begin{minipage}[t]{0.5\textwidth}
  \centering
  \begin{tabular*}{\linewidth}{@{\extracolsep{\fill}}lccccc}
    \toprule
    \multicolumn{6}{c}{(a)~Ablation on \qthreesix{}} \\
    \midrule
    \textbf{Prec.} & BF16 & \textbf{Greedy} & \textbf{ASHA} & \textbf{w/o infra.} & \textbf{\sys{}} \\
    \midrule
    \wfa{}  & 79.8 & \underline{78.9} & \underline{78.9} & 78.8 & \textbf{79.0} \\
    \wfae{} & 79.8 & 76.5 & 76.3 & \underline{77.0} & \textbf{77.5} \\
    \wta{}  & 79.8 & 75.0 & \underline{75.5} & \underline{75.5} & \textbf{75.7} \\
    \bottomrule
  \end{tabular*}
  \end{minipage}\hfill
  \begin{minipage}[t]{0.45\textwidth}
  \centering
  \begin{tabular}{llccc}
    \toprule
    \multicolumn{5}{c}{(b)~A fixed recipe} \\
    \midrule
    \textbf{Model} & \textbf{Prec.} & \textbf{Best public} & \textbf{Fixed recipe} & \textbf{\sys{}} \\
    \midrule
    Qwen3.5  & \wfa{} & \underline{71.8} & \underline{71.8} & \textbf{72.1} \\
    Qwen3.6  & \wfa{} & 77.7 & \underline{78.6} & \textbf{79.0} \\
    Qwen3.8  & \wfa{} & 78.3 & \underline{78.4} & \textbf{79.8} \\
    R1-Llama & \wfa{} & 43.3 & \underline{43.7} & \textbf{44.1} \\
    \midrule
    Qwen3.5  & \wfae{} & 67.1 & \underline{70.1} & \textbf{72.3} \\
    Qwen3.5  & \wta{}  & \underline{65.2} & 61.8 & \textbf{66.8} \\
    \bottomrule
  \end{tabular}
  \end{minipage}

  \label{tab:fixed}
\end{table*}

\MyPara{Scaling to a 320B model and \nvfp{}.}
\label{sec:eval:new}
NVIDIA~\citeyearpar{nvidia2026glmnvfp4} and Red Hat AI~\citeyearpar{redhat2026glmnvfp4} have each released an \nvfp{} checkpoint of \glmflash{}~\cite{zai2026glm53flash}.
\sys{} recovers 98.3\% of the FP8 accuracy on average, against 96.9\% and 96.3\% for the two releases (\Cref{tab:glm}), and it is the most accurate of the three on every benchmark except GPQA-Diamond, where it is 0.1 points below Red Hat AI.
It also quantizes more modules than either release, so its checkpoint is the smallest, at 183.6\,GiB against 190.4 and 184.3\,GiB.
We run one search on this 288-expert model from its FP8 release, with the same agent, library and evaluator as the main searches and no public recipe.
It takes about 6.5 hours on 16 GB300 GPUs and \$69 of agent tokens.
All three \nvfp{} checkpoints are evaluated with the same engine, GPUs and sampling settings (App.~\ref{app:protocol}).

\MyPara{Efficiency.}
\label{sec:eval:efficiency}
\sys{} adds no inference overhead, since every checkpoint keeps the format and vLLM kernel of GPTQ and AWQ at its precision, or the standard \nvfp{} format on \glmflash{}.
In contrast, ParoQuant adds an online rotation that needs an adapted kernel for each format and engine, so \sys{} decodes 1.30--1.75$\times$ faster at \wfa{} (\Cref{tab:efficiency}).
Moreover, \sys{}'s answers are 4\% shorter than the strongest public recipe's over the 12 pairs, and 1.7 points fewer hit the length limit.

\subsection{Ablation Study}
\label{sec:eval:ablation}
\Cref{tab:ablation} replaces one component of \sys{} at a time on \qthreesix{}, where builds cost the most.

\MyPara{Replacing the agent.}
To test whether the agent matters, we replace it with two search algorithms, greedy search and ASHA~\cite{li2020asha}, that search the same library (App.~\ref{app:prompts}).
\sys{} beats both by 0.2--1.2 points at \wfae{} and \wta{}, and they come within 0.1 points at \wfa{}.

\MyPara{Replacing the infrastructure.}
To test our library and harness, we give the agent public libraries and a plain GPU queue, and it writes the rest of the code itself (App.~\ref{app:prompts}).
Over three searches, \sys{} is 0.2--0.5 points more accurate on average and varies less (standard deviation 0.1--0.2 against 0.2--0.4 points).
Without the library, the agent rebuilds parts of it, such as an FP8-aware GPTQ, and writes three times as much code (3.1k against 1.1k lines on average).
It never tries a rotation, the step behind \sys{}'s largest gap at \wfae{}.
It launches 50\% more jobs and loses twice the GPU-hours to failed or cancelled ones (6.6\% against 3.1\%).

\MyPara{A fixed recipe.}
Five of \sys{}'s 12 final recipes follow the same pattern: GPTQ calibrated on self-generated data, then scale distillation.
To test this pattern alone, we fix its settings and apply it as a fixed recipe without a search (\Cref{tab:ablation}b, App.~\ref{app:baselines}).
This pattern was found by \sys{}'s searches, so the fixed recipe is not an independent baseline.
At \wfa{}, it matches or beats the best public recipe on all four models, yet stays below \sys{} on each.
On \qthreefive{}, it is 2.2 and 5.0 points below \sys{} at \wfae{} and \wta{}, so even this fixed recipe does not carry over.

\subsection{What the Searches Found}
\label{sec:eval:case}
\label{sec:eval:traces}
\label{sec:eval:cost}

The 12 final recipes differ in every stage (\Cref{fig:eval-recipes}), and the search logs show two reasons.

\MyPara{No published practice helps every model and precision.}
As \Cref{fig:search-findings}(a) shows, each of four published practices that the searches tried lowered the KL on one model and not on another.
At \wfa{}, calibrating on self-generated data~\cite{williams2025selfcalib} lowers the KL of \qthreesix{} by 30\% but raises it on R1-Llama, whereas scale distillation on the default calibration text helps R1-Llama and not \qthreesix{}.
AWQ's scaling cuts the KL of \qthreefive{} by a third but adds almost nothing after a Hadamard rotation, which treats the same outlier channels.
A plausible reason is that each practice fixes one source of error and helps only where that source dominates.
Across whole recipes, the step that lowered the KL most also differs, a rotation in five searches, distillation in four, self-generated data in two and an FP8 scale search in one (\Cref{fig:search-findings}(b)).

\MyPara{The agent builds its own data and code.}
Every final recipe calibrates or distills on self-generated data for prompts that the agent chose.
In addition, 10 of the 12 train on answers that the quantized model itself wrote.
Five also keep code that the agent wrote for training or merging.
For example, with the codes fixed, the agent's distillation also trains the parameters that the format leaves in BF16, such as normalization gains and the gate and convolution parameters of Gated DeltaNet, and none of them adds inference cost.
In five of the ten, the agent added this training right after a generation check found loops or cut-off answers (\refsec{sec:design:doubt}).
It also diagnosed problems from its logs (bold notes in \Cref{fig:search-findings}(b)).
On \qthreefive{} at \wfa{}, it traced spikes in its distillation loss to BF16 scales that crossed a rounding boundary all at once, which made Adam's steps jump, and raised Adam's $\epsilon$.
On \leight{} at \wta{}, it found that its own rounding code gave the scales a zero gradient and fixed it.
On \qthreeeight{} at \wfae{}, it skipped distillation, because the kernel stores each scale in FP8 and the smallest change to an FP8 scale rescales its whole group of weights by at least 6\%, too large a step to fine-tune.

\section{Related Work}
\label{sec:related}
\MyPara{Quantization of LLMs.}
Weight-only methods~\cite{frantar2023gptq,lin2024awq,dettmers2024spqr,tseng2024quipsharp,tseng2024qtip,egiazarian2024aqlm,paroquant2026,dadgarnia2026gsq} speed up memory-bound decoding.
Weight-activation methods~\cite{xiao2023smoothquant,ashkboos2024quarot,liu2025spinquant,lin2024duquant,zhao2024atom,lin2025qserve} also run the matrix multiplications in low precision, and recent work extends quantization to 4-bit microscaling formats~\cite{rouhani2023microscaling,nvidia2025nvfp4} and to the recurrent states of linear attention~\cite{pan2026leapquant}.
Benchmarks~\cite{gong2024llmc} and large-scale studies~\cite{kurtic2025bf16,liu2025quanthurts,lian2026tokeninflation} compare fixed recipes to distill best practices.
The calibration data~\cite{williams2024calibration,williams2025selfcalib} and tuning against the original model~\cite{shao2024omniquant,chen2025efficientqat,malinovskii2024pvtuning} change the result further, and the KL divergence from the original model is a common way to compare quantized checkpoints~\cite{dutta2024accuracy,llamacpp2024kld}.
\sys{} searches for the combination of these methods, settings and data for each model and precision.

\MyPara{Automated quantization search.}
AMC~\cite{he2018amc}, HAQ~\cite{wang2019haq} and APQ~\cite{wang2020apq} searched per-layer pruning ratios, bit-widths and architectures of CNNs, with the hardware in the loop.
For LLMs, AMQ~\cite{lee2025amq} and EvoPress~\cite{sieberling2025evopress} search per-layer bit-widths, OptiKIT~\cite{optikit2026} runs a user-chosen recipe on several calibration samples and tunes serving parameters, and HAQA~\cite{deng2026haqa} tunes fine-tuning hyperparameters and kernel settings with an LLM agent, but none of them searches over combinations of PTQ methods, their settings and calibration data at a fixed precision.
ModelOpt's recipe-search skill~\cite{nvidia2026modeloptskill} has a coding agent choose among ModelOpt's built-in formats, calibration algorithms and quantized modules, and it writes no new methods or data.
EvoPress also scores by KL, but on models stitched from precomputed layers, not on the served checkpoint.

\MyPara{LLM agents as optimizers.}
LLM agents optimize prompts~\cite{yang2024opro}, algorithms~\cite{romera2024funsearch,novikov2025alphaevolve}, post-training recipes~\cite{liu2026sro}, GPU kernels~\cite{ouyang2025kernelbench,baronio2025kevin,cao2026ksearch,flashinferbench2026} and mechanisms of computer systems~\cite{adrs2025,hamadanian2025glia,zhao2026whoatemyflops}.
ML engineering and research agents follow a validation score~\cite{chan2024mlebench,toledo2025aira} and can overfit or game it~\cite{metr2025rewardhacking,zhong2026impossiblebench}, which AIRA$_2$~\cite{hambardzumyan2026aira2} counters by hiding its search split from the agent.
\sys{} instead scores candidates in minutes with an agent-invisible KL evaluator and evaluates only the final checkpoint on the test benchmarks.

\section{Conclusion}
\label{sec:conclusion}
We present \sys{}, a system in which an LLM agent designs the quantization recipe for each model and precision, because the best recipe differs across them.
\sys{} scores each candidate in minutes by its output KL divergence from the original model on reference traces that the agent never sees, checks a few promising candidates with long generations, and builds them with a library of public PTQ methods that compose and run on new architectures.
On four reasoning models, it improves average accuracy over the strongest public recipe by \textbf{0.9--2.2} points. Given ModelOpt's workflow and tools instead, the same agent falls 0.5--1.3 points behind.
On the 320B-parameter \glmflash{} at \nvfp{}, it recovers 98.3\% of the FP8 accuracy, against 96.9\% for the best vendor checkpoint. Every \sys{} checkpoint adds no inference overhead.

\section*{Acknowledgements}
We thank Han Cai, Yi Pan, Haocheng Xi, Yujun Lin and Ziyi Xu for their insightful feedback.
We thank SK Telecom and the MIT Generative AI Impact Consortium (MGAIC) for supporting this research.
We thank Jenia Jitsev and LAION for providing access to computing resources.
We gratefully acknowledge the Gauss Centre for Supercomputing e.V.\ (GCS) for computing time on the supercomputer JUPITER at the J\"ulich Supercomputing Centre (JSC).

\bibliographystyle{mlsys2025}
\bibliography{refs}

\clearpage
\appendix
\section{Baselines}
\label{app:baselines}

\MyPara{Defaults.}
GPTQ uses llm-compressor's \texttt{GPTQModifier} with its preset for each precision.
The preset is symmetric with $g=128$, a damping fraction of 0.01, static activation order and a min-max observer.
At \wfa{}, \Cref{tab:main} reports the asymmetric variant.
AWQ uses \texttt{AWQModifier} with its defaults (duo scaling, a grid of 20) followed by round-to-nearest, asymmetric at \wfa{} and symmetric at \wta{}.
\wfae{} uses FP8 (E4M3) activations with dynamic per-token scales.
Within one model and precision, every row of \Cref{tab:main} quantizes the same modules at the same bit width.
At \wta{}, the output head and the token embeddings are INT4 with $g=64$ in every row, as in GSQ's release.

\MyPara{ParoQuant and GSQ.}
ParoQuant uses the checkpoints released by its authors, served on vLLM~0.27.1 with its official rotation kernel.
Its vLLM backend had no MoE support, so we added it for \qthreesix{} with 176 lines of code.
GSQ uses the released checkpoint on \qthreeeight{} and the official code and configuration on \qthreefive{}, \qthreesix{} and \leight{}.
On \qthreesix{}, the official code needed six patches of about 220 lines in total.
Its MoE path skipped the attention layers, so we ported the attention stage of its dense path to quantize attention as in the other rows.
The other patches fixed layer names and a compressed-tensors call that no longer matched the model and the library.
On \leight{}, we use the configuration that its README gives for Llama models.

\MyPara{QuaRot+GPTQ.}
We fuse QuaRot's Hadamard rotations~\cite{ashkboos2024quarot} into the BF16 weights offline and then run the default GPTQ recipe of the \wfae{} row unchanged, with the same calibration data.
On \leight{}, we apply R1 to the residual stream and R2 to the value and output projections of each attention head.
On the Qwen models, we apply R1 only, because the gate before each attention output projection and the Gated DeltaNet layers make R2 inexact.
QuaRot's online rotations (R3 and R4) would add inference operators, which the formats of \Cref{tab:precisions} do not allow, so we leave them out.

\MyPara{FP8-aware GPTQ and the fixed recipe.}
FP8-aware GPTQ (\Cref{fig:motivation}(b), \refsec{sec:motivation:transfer}) is built from scratch with the default calibration data on each model, and its \wfae{} averages are 62.5, 76.4, 76.7 and 42.7 on \qthreefive{}, \qthreesix{}, \qthreeeight{} and \leight{}.
The fixed recipe of \Cref{tab:ablation}b is the pattern that five \sys{} searches found (\refsec{sec:eval:ablation}). It runs GPTQ on self-generated data, a pool of 640 answers that the BF16 model wrote (2{,}048 samples of 2{,}048 tokens).
It then runs two rounds of 200 steps of scale distillation on the same pool, at learning rates $3\times10^{-4}$ and $10^{-4}$, from scratch on every pair.

\section{Evaluation Protocol}
\begin{table}[!htb]
  \centering
  \caption{\textbf{Models in the main suite.} Hybrid models interleave full-attention layers with linear-attention (Gated DeltaNet) layers.}
  \label{tab:models}
  \vskip 0.08in
  \footnotesize
  \resizebox{\columnwidth}{!}{
  \begin{tabular}{lll}
    \toprule
    \textbf{Model} & \textbf{Attention} & \textbf{MLP} \\
    \midrule
    \qthreefive{}  & hybrid, Gated DeltaNet & dense \\
    \qthreesix{}   & hybrid, Gated DeltaNet & MoE, 8 of 256 \\
    \qthreeeight{} & hybrid, Gated DeltaNet & dense \\
    \leight{}      & full & dense \\
    \bottomrule
  \end{tabular}
  }
\end{table}

\label{app:protocol}

\MyPara{Quantized modules.}
We quantize the same modules as the public baselines, namely every decoder linear layer except the two small gate projections of each DeltaNet layer, which most public checkpoints of these models also keep in BF16.
On \qthreesix{}, we also keep the router, the shared expert and its gate in BF16, as in ParoQuant's release.

\MyPara{Main suite.}
Math has 63 problems (AIME~2025 and HMMT~Feb~2026) with 32 samples each, and GPQA-Diamond has 198 questions with 16.
LiveCodeBench~v6 has 454 problems with 4, IFBench 300 prompts with 16 and IFEval 541 prompts with 16.
IFEval is scored prompt-level strict and IFBench prompt-level loose.
Generation stops at 81{,}920 tokens on \qthreefive{} and \qthreesix{}.
On \qthreeeight{}, it stops at 81{,}920 for math and code and at 32{,}768 for GPQA and IFBench, which cuts off only 3\% of its GPQA answers.
On \leight{}, it stops at 32{,}768, the limit that its model card sets.
Each checkpoint is served by vLLM~0.27.1, one engine per GPU on four GPUs, and sample $j$ of every problem uses seed $17+j$.
The answer is the text after the first closing thinking tag.
Math and GPQA answers are read by a strict extractor, and by a lenient one only if the strict one finds nothing.
LiveCodeBench runs every test in a sandbox without network, with an 8-second limit per case.
On \leight{}, GPQA-Diamond uses the simple-evals prompt that DeepSeek-R1 reports~\cite{deepseekai2025r1}, because its answers to the default prompt fail to parse in 4.9\% of samples.

\MyPara{\glmflash{}.}
All four columns of \Cref{tab:glm} are served by vLLM~0.31 on four GB300 GPUs with tensor parallelism 4, temperature 1.0, top-$p$ 0.95 and the same seeds for every model.
Generations stop at 65{,}536 tokens for math and LiveCodeBench and at 32{,}768 otherwise.
Math pools AIME~2026, HMMT~Feb~2026 and HMMT~Nov~2025 (93 problems).
LiveCodeBench~v6 covers 2024-08 to 2025-05, and SuperGPQA uses three subsets of 3{,}000 questions.
NVIDIA's checkpoint is its second release (revision 1d00738).
Unlike Red Hat AI's checkpoint, ours also quantizes the dense MLPs of layers 0--2.
Unlike NVIDIA's, it quantizes the multi-token-prediction head to FP8.

\section{Data, Contamination and the Evaluator}
\label{app:calibration}

\MyPara{Calibration data.}
The baselines are calibrated on 2{,}048 texts (0.91--0.96M tokens), 1{,}536 from llm-compressor's default mix and 512 from UltraChat.
No 20-gram of these texts occurs in any test benchmark.
The agent may build its own data only from allow-listed public datasets, from which every row that holds a test item is removed by 20-gram matching before the search.
Every file that the agent's container can read is then scanned against the test items, with one positive control per benchmark, and a single hit stops the search.
Every search in this paper passed with no hit.

\MyPara{The reference traces.}
The reference traces are two BF16 continuations for each of 48 prompts from AIME 2000--2020, SuperGPQA, OpenThoughts and OpenHermes.
They are checked against every test benchmark and the calibration data by id, normalized-text hash and 13-gram matching, with no hit.
The noise $\sigma$ is the pooled within-recipe standard deviation of $\ln \mathrm{KL}$ over two \wfa{} recipes, each rebuilt on three random subsets of the default calibration file.
It is 0.0159 on \qthreefive{}, 0.0047 on \qthreesix{}, 0.0093 on \qthreeeight{} and 0.0063 on \leight{}.
Scoring one checkpoint twice gives identical values, and two nodes that score the same checkpoint agree within 0.2\%.

\MyPara{Signal comparison.}
The 36 checkpoints of \Cref{fig:motivation}(c) are 31 public baselines of \Cref{tab:main}, a symmetric GPTQ on \qthreeeight{} at \wfa{}, and four scripted recipes on \leight{} at \wfa{}. The 31 baselines are all those of \Cref{tab:main} except the four QuaRot+GPTQ rows and GSQ on \leight{}, which were evaluated later.
The four recipes are RTN, symmetric GPTQ, AWQ followed by GPTQ, and GPTQ calibrated on the original model's outputs, and every checkpoint is evaluated with the protocol of \Cref{tab:main}.
Pairs within a model and precision whose accuracies differ by less than 0.05 points are dropped, which leaves 44.
The layer-wise error is the median relative output error of the linear layers on BF16 inputs.

\MyPara{Cost of a score.}
On one GH200, a KL score takes 3.2--4.0 GPU-minutes on \leight{}, 4.7--6.1 on \qthreefive{}, 5.6--7.6 on \qthreesix{} and 8.0--11.7 on \qthreeeight{}.
One full evaluation takes 23.3--26.0, 74.7--82.1, 59.9--86.2 and 65.1--71.7 GPU-hours on the same models.
The reference traces of our searches cover 0.49--1.13 million positions.
Their top-$k$ cache takes 0.47--1.09\,GiB, whereas full BF16 distributions would take 0.21--0.56\,TB, and the bucket of \Cref{eq:kltopk} holds on average 0.015--0.048\% of the original model's probability.
On nine checkpoints of \qthreeeight{} and \leight{} that we scored both with the exact KL over the full vocabulary and with the top-$k$ estimate, the estimate is 0--0.4\% lower.

\section{Library, Recipes and Efficiency}
\label{app:library}
\label{app:components}
\label{app:efficiency}
\begin{figure*}[t]
    \centering
    \includegraphics[width=\textwidth]{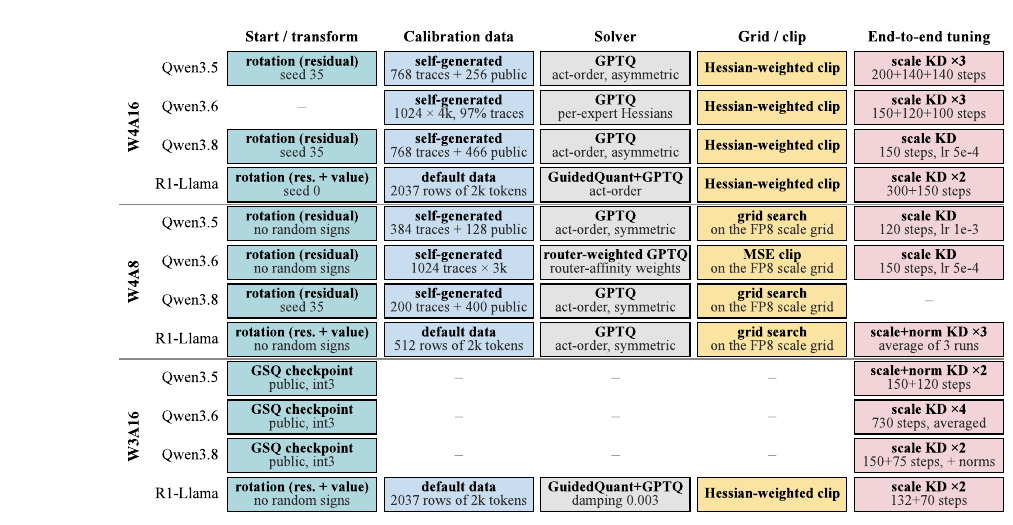}
    \caption{\textbf{Final recipes of the 12 searches.} Each row is one final checkpoint, read left to right. ``Self-generated'' calibration mixes reasoning traces that the BF16 model wrote with public text, scale KD distills the BF16 model into the quantized one by tuning the scales (and, where marked, the norms), and $\times n$ marks $n$ rounds of it.}
    \label{fig:eval-recipes}
\end{figure*}

\Cref{tab:library} lists the methods of the library and the architectures that each one handles, and \Cref{fig:eval-recipes} shows the final recipe of every search.
The library derives the role of every layer from the model, and a model-specific adapter covers only what it cannot derive.
The adapter has no lines for \leight{}, \qthreefive{} and \qthreeeight{}, 14 lines for the fused experts of \qthreesix{} and 72 lines for \glmflash{}, whose FP8 release and multi-token-prediction layer also required extending the library.
\Cref{tab:efficiency} measures decoding on \qthreeeight{} at \wfa{}, where ParoQuant rotates the input of every linear layer online before the same Marlin GEMM as \sys{}.
Each run uses CUDA graphs and 1{,}024 input and 4{,}096 output tokens per request, and we report the median of three runs.
\begin{table}[!ht]
  \centering
  \caption{\textbf{Decode throughput} (output tokens/s) of the \wfa{} checkpoints of \qthreeeight{}, measured together on one GH200 with vLLM~0.27.1. \sys{} runs on the same kernel as GPTQ and decodes within 0.2\% of it, whereas ParoQuant adds an online rotation. Speedup is \sys{} over ParoQuant.}
  \label{tab:efficiency}
  \vskip 0.08in
  \footnotesize
  \begin{tabular}{lcccc}
    \toprule
    \textbf{Batch} & ParoQuant & GPTQ & \textbf{\sys{}} & Speedup \\
    \midrule
    1 & 60 & 105 & 105 & 1.75$\times$ \\
    16 & 805 & 1{,}262 & 1{,}263 & 1.57$\times$ \\
    64 & 1{,}789 & 2{,}330 & 2{,}333 & 1.30$\times$ \\
    \bottomrule
  \end{tabular}
\end{table}

\begin{table}[t]
  \centering
  \caption{\textbf{The 21 methods of the library}, grouped by the decisions of \refsec{sec:background:techniques} (three clipping searches share a row), and the architectures that their code handles: full-attention (\leight{}), Gated DeltaNet hybrid (\qthreefive{}, \qthreeeight{}) and hybrid MoE (\qthreesix{}).}
  \label{tab:library}
  \vskip 0.08in
  \footnotesize
  \resizebox{\columnwidth}{!}{
  \begin{tabular}{llccc}
    \toprule
    \textbf{Method} & \textbf{Source} & \textbf{Full-attn.} & \textbf{Hybrid} & \textbf{Hybrid MoE} \\
    \midrule
    \multicolumn{5}{l}{\textit{Transformations}} \\
    Hadamard rotation (residual) & QuaRot & \checkmark & \checkmark & \checkmark \\
    Hadamard rotation (value heads) & QuaRot & \checkmark & -- & -- \\
    Channel permutation & deepcompressor & \checkmark & \checkmark & \checkmark \\
    Activation-aware scaling & AWQ & \checkmark & \checkmark & \checkmark \\
    \multicolumn{5}{l}{\textit{Rounding}} \\
    Round-to-nearest & -- & \checkmark & \checkmark & \checkmark \\
    GPTQ & GPTQ & \checkmark & \checkmark & \checkmark \\
    GPTAQ & GPTAQ & \checkmark & \checkmark & dense part \\
    Learned rounding & GSQ & \checkmark & \checkmark & \checkmark \\
    Sign gradient rounding & AutoRound & \checkmark & \checkmark & \checkmark \\
    Output-gradient token weights & GuidedQuant & \checkmark & \checkmark & \checkmark \\
    Fisher-trace token weights & ours & \checkmark & \checkmark & \checkmark \\
    Router-weighted expert Hessian & MoEQuant & -- & -- & \checkmark \\
    \multicolumn{5}{l}{\textit{Scales and end-to-end tuning}} \\
    Min-max, MSE, importance-weighted MSE & llm-compressor & \checkmark & \checkmark & \checkmark \\
    Hessian-weighted MSE & ours & \checkmark & \checkmark & \checkmark \\
    FP8 scale grid (\wfae{}) & CUTLASS W4A8 & \checkmark & \checkmark & \checkmark \\
    Scale-only distillation & KD & \checkmark & \checkmark & \checkmark \\
    \multicolumn{5}{l}{\textit{Calibration data}} \\
    Calibration mixing and self-generation & -- & \checkmark & \checkmark & \checkmark \\
    \multicolumn{5}{l}{\textit{Composition}} \\
    Module splicing & -- & \checkmark & \checkmark & \checkmark \\
    Head and embedding format & GSQ release & \checkmark & \checkmark & \checkmark \\
    \bottomrule
  \end{tabular}
  }
  \par\smallskip{\scriptsize The value-head rotation needs a value path without an output gate, which the hybrid models lack.}
\end{table}

\section{Agent, Task Statement and Ablation Variants}
\label{app:prompts}
\label{app:traces}
\label{app:runtime}

\MyPara{Agent.}
Every search runs Claude Code v2.1.303 with \texttt{claude-opus-5-5} (1M-token context, high effort, extended thinking).
No search filled the context window, so Claude Code never had to summarize an agent's context.

\MyPara{Variants.}
The variant of \Cref{tab:ablation} without the infrastructure receives the task statement without the lines about the library, has none of the library's development prompts and answer checkers, and submits GPU jobs to a plain queue with four commands (add, list, log, cancel) instead of the full scheduler.
It keeps the method cards, the evaluator, the submission rule and the default calibration file.
The ModelOpt variant runs one search per precision with ModelOpt at commit 1ca8bd1 and its quant-recipe-search and ptq skills, and it has no access to our library or evaluator.
Its task statement gives the three inputs that the skill asks for: accuracy at a fixed format, the format itself and the four test benchmarks, which a tool runs at fewer samples than the final evaluation.
ModelOpt's exporter refuses 3-bit weights and the MoE block of \qthreesix{}, so we provide a converter that stores ModelOpt's integer codes and scales unchanged in the format of each precision.
\begin{figure}[t]
    \centering
    \includegraphics[width=\columnwidth]{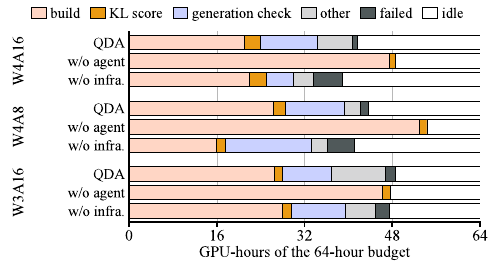}
    \vspace{-14pt}
    \caption{\textbf{Where the 64 GPU-hours of each ablation search go} (\qthreesix{}). ``w/o agent'' is greedy search of \Cref{tab:ablation} and ``w/o infra.'' is the agent without the infrastructure. ``Generation check'' is the agent's checks of its own checkpoints, ``other'' is mostly calibration and training data that the agent generates, and ``failed'' is every job that errored, ran out of memory or was cancelled.}
    \label{fig:eval-hours}
\end{figure}

\Cref{fig:eval-hours} splits the budget of the searches of \Cref{tab:ablation} by the program that each job runs.

\begin{table}[!htb]
\centering
  \caption{\textbf{Search space of greedy search and ASHA} on \qthreesix{} at \wfa{}, taken from the library's method table. Steps whose method does not support this precision or architecture are left out.}
\label{tab:greedy}
  \vskip 0.08in
\footnotesize
\begin{tabular}{@{}p{1.6cm}p{6.2cm}@{}}
\toprule
\textbf{Slot} & \textbf{Steps (grids)} \\
\midrule
Start & BF16 model; public AWQ and GPTQ checkpoints \\
Transform & AWQ scale search writing a new BF16 base; channel permutation; Hadamard rotation of the residual stream \\
Solver & RTN, two implementations (min-max, MSE, importance-weighted MSE); GPTQ (min-max, MSE, Hessian-weighted MSE, importance-weighted MSE); GPTQ with router-weighted expert Hessians (same four); AWQ then RTN (min-max, MSE, importance-weighted MSE); MSE and importance-weighted clip search with RTN; AutoRound (min-max) \\
Refinement & End-to-end scale tuning by distillation from the BF16 model (codes and zero points frozen) \\
\bottomrule
\end{tabular}
\end{table}

\MyPara{Predefined searches.}
The greedy search and ASHA use no LLM.
A recipe takes up to two transforms, one solver and up to two refinements from the library's method table (\Cref{tab:greedy}), or one or two refinements of a public checkpoint.
It also chooses its calibration data, which is the default file, a fixed pool of 640 answers that the original model wrote before the search, or a mix of the two.
Finally, it chooses 512 or 2{,}048 samples of 2{,}048 or 4{,}096 tokens and the maximum length and learning rate of distillation.
The greedy search starts from the public checkpoint with the lowest KL and repeatedly tries, at random, an untried recipe one edit away from the best one so far.
ASHA samples recipes that start from scratch or from the lowest-KL public checkpoint and runs successive halving over distillation steps, with rungs at 23, 67 and 200 steps and the top third promoted.
Both submit their three lowest-KL checkpoints, and the GPU time to generate the pool counts against their budget.
The answers in the pool run to at most 12{,}288 tokens.

\MyPara{The task statement.}
Each search starts with one instruction, which tells the agent to read the task statement in full, follow it and keep the best submitted checkpoint until the deadline.
\Cref{tab:task} summarizes the statement for \qthreesix{} at \wfa{}.
It names no method to use or to avoid, and its full text ships with the artifact.
Across models, the statement changes in the model name, the default calibration file, the clause on the expert kernel and, on \leight{}, the length of the long-generation checks (32{,}768 tokens).
Across precisions, it also changes in the goal sentence, the list of baselines and the format section.
\begin{table}[t]
\centering
  \caption{\textbf{The task statement} for \qthreesix{} at \wfa{}, abridged. Every variant of \Cref{tab:ablation} receives the same text, except that the variant without the infrastructure receives it without the lines about the library.}
\label{tab:task}
  \vskip 0.08in
\footnotesize
\begin{tabular}{@{}p{1.55cm}p{6.25cm}@{}}
\toprule
\textbf{Section} & \textbf{What it says} \\
\midrule
Goal & The best \wfa{} checkpoint by the deadline. It is scored later on math, code, science and instruction following, which count equally, and nothing measurable in the workspace is that evaluation. \\
Clock and resources & Two nodes of four GH200s until the deadline, when every task is cancelled. Plan for idle GPUs in the last hour. Run long-generation checks (up to 81,920 tokens) on the top checkpoints in all four domains before submitting. Training data may include the quantized model's own answers scored by the BF16 model. When the latest checkpoints of the current line of recipes tie in KL, try another line. Reread the notes and failed logs every couple of hours. \\
Workspace and tools & The quantization and serving environments, offline datasets, a format-compatibility package with no quantization method, and the \texttt{\syscmd} library: atomic transforms, solvers and refinements to stack, plus measurements (\texttt{proxy}, \texttt{profiler}), diagnostics (\texttt{probe}) and \texttt{mix}. The method cards, loaded as a Claude Code skill. New methods and subagents are allowed. \\
Prior & The BF16 model and the public checkpoints with their methods. ParoQuant is listed for reference only, since it needs extra kernels. A public checkpoint cannot be submitted unchanged. \\
Format & INT4, $g=128$, asymmetric, on the Marlin kernels; the same modules and bits as the public checkpoints; nothing added at inference. Forbidden: other bit widths or per-module bits, modules from foreign checkpoints, mixed symmetric and asymmetric modules. \\
Data & No build data may contain an evaluation item or a text of the reference traces. \\
KL & \texttt{kl DIR} checks a checkpoint and returns its KL from the BF16 model, in total, per domain and on human-written text. \texttt{kl log} lists past KL values, and \texttt{kl sens} estimates which module groups would close a KL gap. The reference traces are hidden. \\
Submission & \texttt{handin submit} takes a ranked list; the evaluator checks each checkpoint for format and a short inference, not for answers. If no submitted checkpoint passes, the public checkpoint closest to the BF16 model is evaluated. Submitted checkpoints must be written by jobs in the queue. \\
Notes & Keep a report and notes to resume from, and read long outputs from files. \\
Scope & Do not look for or rebuild evaluation items, and do not edit the evaluation, grading, serving or evaluator code. \\
\bottomrule
\end{tabular}
\end{table}

\end{document}